\pdfoutput=1

\documentclass[11pt]{article}

\usepackage[]{acl}
\usepackage[greek,english]{babel}
\usepackage{times}
\usepackage{latexsym}
\usepackage{subcaption}
\usepackage{paralist}

\usepackage{xcolor}

\newcommand{\mycomment}[3]{}

\newcommand{\ignore}[1]{}
\usepackage[T1]{fontenc}

\usepackage[utf8]{inputenc}
\usepackage{amsmath}
\usepackage{array,booktabs}
\usepackage{graphicx}
\usepackage{textcomp}
\usepackage{booktabs}
\usepackage{colortbl}
\usepackage{amssymb}
\definecolor{columbiablue}{rgb}{0.61, 0.87, 1.0}
\definecolor{cinnabar}{rgb}{0.89, 0.26, 0.2}
\newcommand{\n}{\cellcolor{cinnabar!15}}
\newcommand{\p}{\cellcolor{columbiablue!25}}
\newcommand{\enoc}{\textbf{\textsc{en}}$\mathbf{\rightarrow}$\textbf{\textsc{oc}}}
\newcommand{\itoc}{\textbf{\textsc{it}}$\mathbf{\rightarrow}$\textbf{\textsc{oc}}}
\newcommand{\enbe}{\textbf{\textsc{en}}$\mathbf{\rightarrow}$\textbf{\textsc{be}}}
\newcommand{\enmr}{\textbf{\textsc{en}}$\mathbf{\rightarrow}$\textbf{\textsc{mr}}}
\newcommand{\ensw}{\textbf{\textsc{en}}$\mathbf{\rightarrow}$\textbf{\textsc{sw}}}

\newcommand{\ocen}{\textbf{\textsc{oc}}$\mathbf{\rightarrow}$\textbf{\textsc{en}}}
\newcommand{\ocit}{\textbf{\textsc{oc}}$\mathbf{\rightarrow}$\textbf{\textsc{it}}}
\newcommand{\been}{\textbf{\textsc{be}}$\mathbf{\rightarrow}$\textbf{\textsc{en}}}
\newcommand{\mren}{\textbf{\textsc{mr}}$\mathbf{\rightarrow}$\textbf{\textsc{en}}}
\newcommand{\swen}{\textbf{\textsc{sw}}$\mathbf{\rightarrow}$\textbf{\textsc{en}}}
\newcommand{\hz}{\vphantom{\parbox[c]{0.05cm}{\rule{0.05cm}{0.5cm}}}}

\newcommand{\editedside}{%
  \colorbox{red!50}{\tiny{$\rightarrow$}}%
}

\usepackage{microtype}

\title{\textsc{BitextEdit:} Automatic Bitext Editing \\ for Improved Low-Resource Machine Translation}

\author{Eleftheria Briakou$^1$\thanks{~\ Work done during internship at Facebook AI Research.}, \ \   Sida I. Wang$^2$,  Luke Zettlemoyer$^2$, Marjan Ghazvininejad$^2$ \\
  $^1$ University of Maryland,
  $^2$ Facebook AI Research \\
 \texttt{{ebriakou@cs.umd.edu}},
 \texttt{\{sida,lsz,ghazvini\}@fb.com}
 }

\begin{document}
\maketitle

\begin{abstract}
Mined bitexts can contain imperfect translations that yield unreliable training signals for Neural Machine Translation (\textsc{nmt}). While filtering such pairs out is known to improve final model quality, we argue that it is suboptimal in low-resource conditions where even mined data can be limited. In our work, we propose instead, to refine the mined bitexts via automatic editing: given a sentence in a language $\mathbf{x_f}$, and a possibly imperfect translation of it $\mathbf{x_e}$, our model generates a revised version $\mathbf{x_f'}$ or $\mathbf{x_e'}$ that yields a more equivalent translation pair (i.e., <$\mathbf{x_f, x_e'}$> or <$\mathbf{x_f', x_e}$>).
We use a simple editing strategy by (1) mining potentially imperfect translations for each sentence in a given bitext, (2) learning a model to reconstruct the original translations and translate, in a multi-task fashion.
Experiments demonstrate that our approach successfully improves the quality of CCMatrix mined bitext for $5$ low-resource language-pairs and $10$ translation directions by up to $8$ \textsc{bleu} points, in most cases improving upon a competitive translation-based baseline.
\end{abstract}

%
\section{Introduction}\label{sec:introduction}
Neural Machine Translation (\textsc{nmt}) for low-resource languages is challenging due to the scarcity of bitexts, i.e., translated text in two languages~\cite{koehn-knowles-2017-six}. 
Models are often trained on heuristically aligned~\cite{resnik-1999-mining,banon-etal-2020-paracrawl,espla-etal-2019-paracrawl} or automatically mined data~\cite{schwenk-etal-2021-wikimatrix, schwenk-etal-2021-ccmatrix}, which can be low quality~\cite{briakou-carpuat-2020-detecting, mashakane}.
This data can include errors that range from small meaning differences in sentences that overlap in content to major differences that yield
completely incorrect translations and random noise, e.g., empty sequences, text in the wrong language, non-linguistic content, among others. 

\begin{figure}[!t]
    \centering
    \includegraphics[scale=0.38]{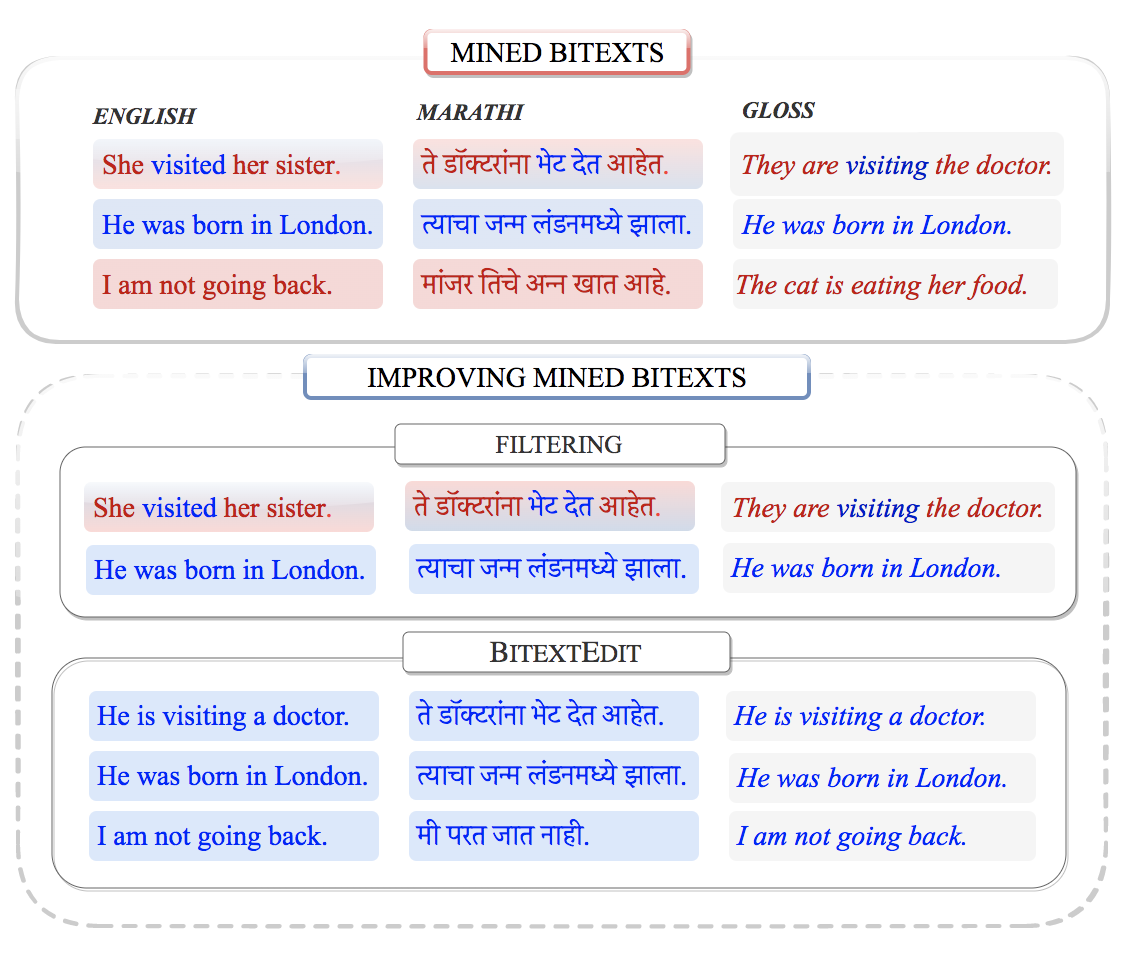}
    \caption{Noisy bitexts consist of a mixture of good-quality, imperfect, and poor-quality translations. Filtering decreases the size of training samples which is crucial for low-resource \textsc{nmt}. Our approach, alternatively, revises noisy bitexts via utilizing imperfect translations in a more effective way, while keeps the size of training data untouched.
    }
    \label{fig:intro_example}\vspace{-0.5cm}
\end{figure}

Filtering out noisy samples from web-crawled bitexts is therefor standard practice for building high quality models~\cite{koehn-etal-2018-findings}, and is particularly helpful in low-resource settings~\cite{koehn-etal-2019-findings,koehn-etal-2020-findings}.  
Despite the popularity of this approach, we argue it has two key limitations. First, partially correct translations provide signal that is lost if the entire example is dropped (see first sample bitext in Figure~\ref{fig:intro_example}). 
Second, filtering out samples exacerbates the data scarcity problem for the long-tail of low-resource language-pairs. 

In this paper, we instead aim to make use of as much of the signal from the mined bitext as possible. We propose an \textbf{editing approach to bitext quality improvement}. Our model takes as input a bitext (i.e., $(\mathbf{x_f,x_e})$), and edits one of the two sentences to generate a refined version of the original (i.e., $\mathbf{x_f'}$ or $\mathbf{x_e'}$) as necessary. By framing the problem as a bitext editing (\textsc{BitextEdit}) task, we can perform a wide range of operations from \textit{copying} good-quality bitext, to \textit{partial editing} of small meaning mismatches, and \textit{translating} from scratch incorrect references. 
Following previous extrinsic evaluations of bitext quality~\cite{koehn-etal-2019-findings, koehn-etal-2020-findings,schwenk-etal-2021-ccmatrix,schwenk-etal-2021-wikimatrix}, we compare \textsc{nmt} models trained on the original and revised versions of CCMatrix bitexts. Concretely, we report consistent improvements in translation quality 
for $10$ low-resource \textsc{nmt} translation tasks:
\textsc{en}$\leftrightarrow$\textsc{oc}, 
\textsc{it}$\leftrightarrow$\textsc{oc}, 
\textsc{en}$\leftrightarrow$\textsc{be}, 
\textsc{en}$\leftrightarrow$\textsc{mr}, 
and
\textsc{en}$\leftrightarrow$\textsc{sw}, while in most cases we even improve  upon a competitive translation-based baseline.
Crucially, \textsc{BitextEdit} yields from $4-8$ \textsc{bleu} point improvements in the more data-scarce settings (i.e., \textsc{en-oc}, \textsc{it-oc}). Additionally, our quantitative and qualitative analyses indicate that \textsc{BitextEdit} improves bitext quality in higher-resource settings with lighter editing that targets more fine-grained meaning differences. 
%
%
\section{Background}\label{sec:background}

\paragraph{Bitext Mining} The idea of using the web as a source of parallel texts has a long history~\cite{resnik-1999-mining}. Recent advances in multilingual representation learning~\cite{artetxe-schwenk-2019-massively, liu-etal-2020-multilingual-denoising} enable the curation of mined bitexts across multiple languages at scale. For instance, combining \textsc{laser}~\cite{artetxe-schwenk-2019-massively} embeddings
with nearest neighbor search allows for effective bitext mining from Wikipedia, i.e., WikiMatrix~\cite{schwenk-etal-2021-wikimatrix} and CommonCrawl monolingual texts, i.e., CCMatrix~\cite{schwenk-etal-2021-ccmatrix}. While the latter approach requires parallel text supervision to train the multilingual sentence representation encoder, 
\citet{NEURIPS2020_1763ea5a} shows that it can be extended to an unsupervised framework via 
iterative self-supervised training. 

\paragraph{Issues in Bitext Quality} ~\citet{mashakane} manually audit the quality of multilingual datasets in $205$ language-specific corpora that result from automatic curation pipelines, including bitexts from CCAligned~\cite{el-kishky-etal-2020-ccaligned}, WikiMatrix~\cite{schwenk-etal-2021-wikimatrix}, and ParaCrawl~\cite{banon-etal-2020-paracrawl,espla-etal-2019-paracrawl}. All have systematic issues, especially for low-resource languages. 
The vast majority of low-resource pairs contain less than $50\%$ valid translations. 
However, they do often share structural similarity and partial content.
\citet{briakou-carpuat-2020-detecting}---in a more fine-grained annotation study---highlight that small content mismatches are even found in high resource pairs: $40\%$ of English-French WikiMatrix sentence-pairs have \textit{small meaning mismatches}. Our work aims at improving bitext quality via eliminating their systematic issues via editing. 

\paragraph{Bitext Quality vs. \textsc{nmt} Training} \citet{khayrallah-koehn-2018-impact} demonstrate the often significant impact of various types of noise on \textsc{nmt}, via increasing the percentage of $5$ types of artificially injected errors on a clean English-German corpus---mimicking frequent issues in parallel texts (i.e., copying, wrong language, non-linguistic content, short segments, empty sequences). 
\citet{pmlr-v80-ott18a} also argue that 
data uncertainty resulting from noisy references contributes to the miscalibration of \textsc{nmt} models. 
Apart from noisy references, \textit{small meaning mismatches} have also a measurable impact on various aspects of \textsc{nmt}: \citet{briakou-carpuat-2021-beyond} show that models trained on synthetic divergences output degenerated text more frequently and are less confident in their predictions.
In contrast with prior studies that discuss how imperfect references interact with \textsc{nmt} training \textit{solely} for high-resource pairs, we \textit{primarily} focus on low-resource settings and improve \textsc{nmt} models by improving their training bitexts. 

\paragraph{Bitext Quality Improvement} The most standardized approach to improving bitext 
either discards an example or treats it as a perfect training instance~\cite{koehn-etal-2018-findings}. Past submissions to the Parallel Corpus Filtering \textsc{wmt} shared task employ a diverse set of approaches covering simple pre-filtering rules based on language identifiers and sentence features~\cite{rossenbach-etal-2018-rwth,lu-etal-2018-alibaba,ash-etal-2018-speechmatics}, learning to weight scoring functions based on language models, extracting features from neural translation models and lexical translation probabilities~\cite{sanchez-cartagena-etal-2018-prompsits}, combining pre-trained embeddings~\cite{papavassiliou-etal-2018-ilsp}, 
and dual-cross entropy~\cite{chaudhary-etal-2019-low}.
In contrast to prior work, and similar to ours, \citet{briakou2022}
propose to revise imperfect translations in bitext via selectively replace them with synthetic translations generated by  \textsc{nmt} of sufficient quality.
Our work builds on top of prior work and instead of filtering out all the imperfect bitexts, we selectively edit them and keep them in the pool of training data targeting low-resource \textsc{nmt}.
%
%
\begin{figure*}[t!]
\centering
\includegraphics[width=16cm]{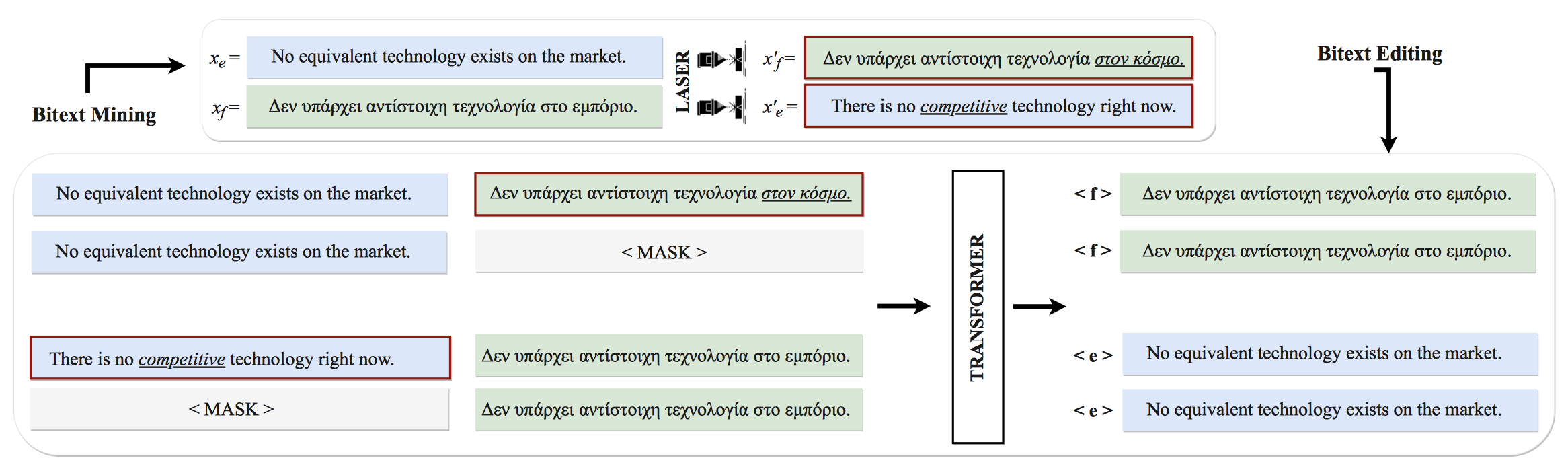}
\caption{\textsc{BitextEdit} training strategy: Our multi-task model is trained using synthetic supervision from mined bitexts. Starting from an original bitext ($x_e, x_f$), we mine imperfect translations $x'_f$ and $x'_e$ for each reference using \textsc{laser} (Bitext Mining). A sequence-to-sequence Transformer model is trained to \textit{translate} and \textit{reconstruct} the original references given synthetically extracted  bitexts representing imperfect translations (Bitext Editing).}\label{fig:approach}
\end{figure*}
\section{Approach: \textsc{BitextEdit}}\label{sec:method}

We frame bitext refinement as an editing task (i.e, \textsc{BitextEdit}) that takes two \textit{input sentences}: a sentence $\mathbf{x_f}$ in language $\mathbf{f}$ and a sentence  $\mathbf{x_e}$ in language $\mathbf{e}$, and aims at editing \textit{one} of them (i.e., it \textit{outputs} $\mathbf{x_f'}$ or $\mathbf{x_e'}$) with the goal of yielding a more equivalent translation pair (i.e., <$\mathbf{x_f}$ , $\mathbf{x_e'}$> or <$\mathbf{x_f'}$, $\mathbf{x_e}$>).
Figure~\ref{fig:approach} gives an overview of our approach while below we describe the bitext refinement model (\S \ref{sec:editing_description}) and the curation of data needed to train our model based on bitext mining (\S \ref{sec:mining_description}). 

\subsection{Bitext Editing}\label{sec:editing_description}

\paragraph{Architecture} Our bitext editing model is a transformer sequence-to-sequence architecture. Each bitext ($\mathbf{x_f}$, $\mathbf{x_e}$) is encoded via adding position embeddings that are reset for each input sentence to facilitate their alignment \cite{NEURIPS2019_c04c19c2} and two language embeddings, initialized at random, to indicate the two languages for the editing model. The decoder generates autoregressively a refined version of $\mathbf{x_f}$ or $\mathbf{x_e}$, where the first generated token indicates which of the two input sentences is edited, as described below.

\paragraph{Learning}  During training, we optimize the multi-task loss presented in Equation~\ref{eq:losss}, which has two components. 
The first represents a \textit{edit-based reconstruction loss} (i.e., $\mathcal{L}_\mathrm{\textsc{edit}}$) that reconstructs one of the two sentences, e.g., $\mathbf{x_f}$ started from a noised version of the original bitexts e.g.,  $\mathbf{x_f'}$ and $\mathbf{x_e}$.
We make this loss bi-directional via adding a symmetrical loss that reconstructs $\mathbf{x_e}$ from $\mathbf{x_f}$ and $\mathbf{x_e'}$, respectively. 
The second component, is implemented as a bi-directional \textit{translation loss} 
(i.e., $\mathcal{L}_\mathrm{\textsc{mt}}$)  
via masking the inputs of the target translation directions (e.g., generate $\mathbf{x_e}$ given $\mathbf{x_f}$ and $\mathrm{\textsc{<mask>}})$. 
Finally, in both losses a language identification symbol (i.e., $\mathrm{\textsc{<}f\textsc{>}}$ or $\mathrm{\textsc{<}e\textsc{>}}$) is used as the initial token to predict the language of the output text.

\begin{equation}
    \tiny
    \begin{aligned}
    \mathcal{L} & = \sum_{(\mathbf{x_f},\mathbf{x_e})} \Bigg( \underbrace{ \mathrm{logp}\Big( [\mathrm{\textsc{<}e\textsc{>}} \ \ \mathbf{x_e}] \mid (\mathbf{x_f}, \mathbf{x_e'})\Big) + \mathrm{logp}\Big( [\mathrm{\textsc{<}f\textsc{>}} \ \ \mathbf{x_f}] \mid (\mathbf{x_f'}, \mathbf{x_e})\Big) }_{\mathcal{L}_\mathrm{\textsc{edit}}}+ \\
                &  \underbrace{\mathrm{logp}\Big([\mathrm{\textsc{<}e\textsc{>}} \ \ \mathbf{x_e}] \mid (\mathbf{x_f}, \mathrm{\textsc{<mask>}})\Big) + \mathrm{logp}\Big([\mathrm{\textsc{<}f\textsc{>}} \ \ \mathbf{x_f}] \mid (\mathrm{\textsc{ <mask>}}, \mathbf{x_e})\Big) }_{\mathcal{L}_\mathrm{\textsc{mt}}} \Bigg)     
    \end{aligned}\label{eq:losss}
\end{equation}

\paragraph{Inference} At test time, our model takes as input a possibly imperfect bitext and edits one of the reference translations, while first generating
the language identification token. The latter is used to infer which of the two reference translations gets revised. Finally, we pair the edited output sequence with the original input that does not get revised, yielding a refined bitext.

\subsection{Bitext Mining}\label{sec:mining_description}
Our model requires access to $\mathbf{x_f'}$ and $\mathbf{x_e'}$ training instances that are treated as noised versions of $\mathbf{x_f}$ and $\mathbf{x_e}$, respectively. Since our goal is to develop a model that can refine mismatches found in mined bitexts at inference time,
we want our noised training instances to share similar properties  with the mined ones, e.g., fluent text in the target language, possibly imperfect translations of the source text.
To this direction, we take a distance-based mining approach to construct the noised samples similar to ~\citet{schwenk-2018-filtering}.  Unlike~\citet{artetxe-schwenk-2019-massively} we do not use a margin score on the \textit{normalized} cosine distance of sentence-pairs to keep the computation cost low and encourage mining of more diverse imperfect translations. 
Concretely, given the mined bitext $(\mathbf{x_f},\mathbf{x_e})$ and two pools of monolingual sentences $\mathcal{F}$ and $\mathcal{E}$, in language $\mathbf{f}$ and $\mathbf{e}$, we extract $\mathbf{x_f'}$ and $\mathbf{x_e'}$ as follows:

\begin{equation}
    \small
    \begin{aligned}
    \mathbf{x_f'} &= \mathrm{argmax}_{\mathbf{z} \in \mathcal{F}} \ \ \mathrm{cos}(\textsc{laser}(\mathbf{z}), \textsc{laser}(\mathbf{x_e}))  \\
     \mathbf{x_e'} &= \mathrm{argmax}_{\mathbf{z} \in \mathcal{E}} \ \ \mathrm{cos}(\textsc{laser}(\mathbf{x_f}), \textsc{laser}(\mathbf{z}))   
    \end{aligned}
\end{equation}

\noindent
where \textsc{laser} \cite{artetxe-schwenk-2019-massively}  represents a multilingual encoder used to extract sentence embeddings for each sentence, while the most similar sentence is returned based on nearest neighbor retrieval. Furthermore, this formula is extended to retrieval of top $k$ sentences, while we also allow mining of the original CCMatrix translations. The latter happens to expose the model to good translations at training time, that should not be edited.
%
%
\section{Experimental Setting}\label{sec:setting}

\paragraph{Bitexts} We focus on CCMatrix data for two main reasons: a) it constitutes the only large-scale available resource for a lot of low-resource language pairs and b) recent efforts of auditing this corpus raise concerns regarding the quality of mined bitext of low-resource pairs. CCMatrix is mined using \textsc{laser} embeddings following the max-strategy approach: a margin score is computed for all monolingual sentences in two languages, then the union of forward and backward candidates is build and pairs that score above a pre-defined threshold are treated as translations. \citet{schwenk-etal-2021-ccmatrix} set the threshold globally for all languages at $1.06$. 

Our primary goal is to explore whether bitexts that are typically discarded by filtering can be refined by our model and thus benefit low-resource \textsc{nmt}. For this purpose, we define two pools of CCMatrix data: Pool~A
corresponds to CCMatrix data with \textsc{laser} scores greater than $1.06$, while Pool~B contains bitexts with scores lower than $1.06$ and greater than $1.05$. The latter threshold is primarily chosen since CCMatrix bitexts is only available above this value. Editing bitexts with even smaller scores is an interesting area for future work. 

\paragraph{Training data} Our models are trained based on procedures described in \S\ref{sec:mining_description}, where we use Pool~A to seed the generation of noised training samples $\mathbf{x_f'}$ and $\mathbf{x_e'}$. We mine $k$ samples $\mathbf{x_f'}$ for each $\mathbf{x_e}$ and $k$ samples $\mathbf{x_e'}$ for each $\mathbf{x_f}$, respectively. We set $k$ to $4$ and include detailed statistics in Appendix ~\ref{sec:artifacts_details}.  

\paragraph{Language-pairs} We experiment with the following languages: English-Occitan (\textsc{en-oc}), Italian-Occitan (\textsc{it-oc}), English-Belarusian (\textsc{en-be}), English-Marathi (\textsc{en-mr}), and English-Swahili (\textsc{en-sw}).
The $5$ language pairs are chosen to include diverse low-resource pairs, which differ either in training data size or language similarity. Table\ref{tab:ccmatrix_stats} summarizes the data conditions.

\begin{table}[!t]
    \centering
    \scalebox{0.8}{
    \begin{tabular}{llcc}
    \rowcolor{gray!10}
    \textbf{\textsc{pair}} & \textbf{\textsc{scripts}}  & Pool~A & Pool~B  \\
    \textsc{en-oc} & Latin-Latin      & $0.2$M  & $0.1$M  \\
    \textsc{it-oc} & Latin-Latin      & $0.3$M  & $0.1$M  \\
    \textsc{en-be} & Latin-Cyrillic   & $0.7$M  & $1.1$M  \\
    \textsc{en-mr} & Latin-Devanagari & $1.5$M  & $2.1$M  \\
    \textsc{en-sw} & Latin-Latin      & $1.7$M  & $0.9$M  \\
    \end{tabular}}
    \caption{Statistics of CCMatrix bitexts.}\vspace{-1em}
    \label{tab:ccmatrix_stats}
\end{table}

\paragraph{Comparisons} We run several extrinsic evaluations using \textsc{nmt} trained on different versions of CCMatrix data. First, we train \textsc{nmt} models on two versions of original CCMatrix data: Pool~\textit{A} \cite{schwenk-etal-2021-ccmatrix} and Pool~$A \cup B$. Second, we aim at revising Pool B via
a) a translation-based approach that revisits the source-side of the bitexts via back-translating their target-side with a model trained on original CCMatrix,
(i.e., $b(.)$)
and b) via editing either the source or the target side of it using our proposed approach (i.e, $r$(.)).

\paragraph{Model details} Our models are implemented on top of \texttt{fairseq} \cite{ott2019fairseq}.\footnote{\url{https://github.com/pytorch/fairseq}} 
We use the same Transformer architecture as in \citet{schwenk-etal-2021-ccmatrix}, with embedding size $512$, $4{,}096$ transformer hidden size, $8$ attention heads, $6$ transformer layers, and dropout $0.4$. We train with $0.2$ label smoothing and Adam optimizer with a batch size of $4{,}000$ tokens per \textsc{gpu}. We include more model details in Appendices~\ref{sec:fairseq_settings} and \ref{sec:infrastructure}. 
We train for $100$ epochs and select best checkpoint based on validation perplexity. We report single run results.

\paragraph{Data Preprocessing} We use the standard Moses scripts \cite{koehn-etal-2007-moses} for tokenization of \textsc{en, oc, it, be} and \textsc{sw} and the Indic \textsc{nlp} library\footnote{\url{https://anoopkunchukuttan.github.io/indic_nlp_library/}} for \textsc{mr}. For each language-pair, we learn $60$K \textsc{bpe}s using \texttt{subword-nmt} \cite{sennrich-etal-2016-neural}.\footnote{\url{https://github.com/rsennrich/subword-nmt}}

\paragraph{Evaluation} We evaluate our models on the \textit{devtest} of \texttt{flores} \cite{guzman-etal-2019-flores}. We report \texttt{spm-bleu}\footnote{\url{https://github.com/facebookresearch/flores}} on detokenized outputs and chrF \cite{popovic-2015-chrf} as our second evaluation metric.\footnote{Results on chrF are included in Appendix~\ref{sec:chrf}.}

\begin{table*}[!t]
    \centering
    \scalebox{0.7}{
    \begin{tabular}{rlr@{\hskip 0.5in}rr@{\hskip 0.5in}rr@{\hskip 0.5in}rr@{\hskip 0.5in}rr@{\hskip 0.5in}rr}
    \toprule
    & & &  \multicolumn{2}{l}{\enoc} & \multicolumn{2}{l}{\itoc} & \multicolumn{2}{l}{\enbe}  & \multicolumn{2}{l}{\enmr} & \multicolumn{2}{l}{\ensw}\\
     \cmidrule(lr){4-13}
    $1:$ & CCMatrix & $A\cup B$  &  $20.5$      &          &   $11.5$  &             & $11.0$           &         & $12.2$           &        &   $38.1$    &  \\
    $2:$ & Filtering         & $A$ &    $18.1$      & \n $-2.4$   &   $11.7$  &  \p $+0.2$     & $9.8$            & \n  $-0.2$ & $12.2$           & $0.0$  &   $37.6$    & \n $-0.5$\\
    
    \addlinespace[0.1cm]

    $3:$ & Translation-based  & $b(A\cup B)$ & $20.8$ & \p $+0.3$  & $17.0$  & \p $+5.5$   &  $12.3$ & \p  $+1.3$ & $15.5$ & \p $+3.2$  & $37.6$ & \n $-0.5$    \\
    $4:$ & \textsc{BitextEdit} & $r(A\cup B)$ & $25.4$ & \p $+4.9$ & $19.8$ & \p $+8.3$ & $12.8$ & \p $+1.7$  & $\mathbf{15.8}$ & \p $+3.6$ & $37.8$ & \n $-0.3$\\
    
    \addlinespace[0.1cm]

    $5:$ & Translation-based  & $A\cup b(B)$ &    $23.0$      & \p $+2.5$   &   $17.0$  &  \p $+5.5$     & $12.1$           & \p  $+1.1$ & $15.4$  & \p  $+3.2$ & $\mathbf{38.8}$ & \p  $+0.7$  \\
    $6:$ & \textsc{BitextEdit} & $A\cup r(B)$ & $\mathbf{26.0}$ & \p $+5.5$   & $\mathbf{19.9}$ & \p $+8.4$ & $\mathbf{13.0}$  & \p $+2.0$ & $15.3$  & \p $+3.1$ & $38.3$ & \p $+0.2$ \\
    
    \addlinespace[0.9cm]
    
    & & & \multicolumn{2}{l}{\ocen} & \multicolumn{2}{l}{\ocit} & \multicolumn{2}{l}{\been}  & \multicolumn{2}{l}{\mren} & \multicolumn{2}{l}{\swen} \\
    \cmidrule(lr){4-13}
    $7:$ & CCMatrix &  $A\cup B$ &   $24.3$       &         &  $11.6$          &          & $9.8$    &            & $13.0$ & & $34.8$ & \\
    $8:$ & Filtering & $A$       &     $17.8$       & \n $-6.5$  & $11.1$           & \n $-0.5$   & $7.8$    & \n $-2.0$    & $11.3$ & \n $-1.07$ & $34.8$   & $0.0$    \\
 
     \addlinespace[0.1cm]
    
    $9:$ & Translation-based  & $b(A\cup B)$ & $26.6$ & \p $+2.3$  & $17.3$ & \p $+5.7$ & $9.9$  & \p $+0.1$ & $13.6$ & \p $+0.6$ & $33.8$ & \n $-1.0$ \\
    $10:$ & \textsc{BitextEdit} & $r(A\cup B)$        & $28.2$ & \p $+3.9$ & $\mathbf{18.5}$  & \p $+6.9$ & $10.7$ & \p $+0.9$ & $16.4$ & \p $+3.4$ & $35.8$ &  \p $+1.1$\\
    
    \addlinespace[0.1cm]
    
    $11:$ & Translation-based  & $A\cup b(B)$ &   $27.7$       & \p $+3.4$  & $15.6$           & \p $+4.0$   & $9.6$    & \n $-0.2$     & $15.1$ & \p $+2.1$ & $\mathbf{36.8}$ & \p $+2.0$\\
    $12:$ & \textsc{BitextEdit} & $A\cup r(B)$ & $\mathbf{28.7}$  & \p $+4.4$  & $18.3$  & \p $+6.7$    & $\mathbf{10.8}$  & \p $+1.0$ &  $\mathbf{16.7}$ & \p $+3.7$ & $36.2$ & \p $+1.8$\\
    \toprule
    \end{tabular}}
    \caption{Results on \textsc{nmt} tasks for models trained on different versions of CCMatrix. For each task the first column denotes spm-\textsc{bleu}; the second columns (highlighted scores) give the difference of each row with the original CCMatrix. Models trained on the refined bitexts improve \textsc{nmt} for low-resource language-pairs.}
    \label{tab:main_results}
\end{table*}

%
%
\section{Experimental Results}\label{sec:results}

\paragraph{Bitext filtering revisited} We first provide empirical evidence that bitext filtering 
might be a suboptimal solution to low-resource \textsc{nmt}.  Table~\ref{tab:main_results} shows that filtering out sentence pairs that score below the predefined threshold of $1.06$ (i.e., Filtering)  surprisingly hurts translation quality in almost all translation tasks (rows $2$ vs. $1$ and $8$ vs. $7$). This result is likely because the threshold was optimized for specific language-pairs, and the fact that---under low-resource regimes---increasing the amounts of \textit{possibly imperfect} translation data might still benefit \textsc{nmt}. Furthermore, this experiment gives us insights on the quality of the training data our bitext editing model uses: for \textsc{it-oc}, \textsc{be-en}, and \textsc{en-mr} we expect Pool~A to provide more noisy training signals (as \textsc{bleu} scores of \textsc{nmt} models trained on it are $\sim 11$), compared to \textsc{en-oc} and \textsc{en-sw} where the quality of the given bitext is expected to be significantly better (\textsc{bleu} scores $\sim 18$ and $\sim 37$, respectively).

\paragraph{Editing Pool~B} 
Applying \textsc{BitextEdit} to edit erroneous translations in Pool~B (i.e., $A \cup r(B)$) improves the quality of \textsc{nmt} systems over the ones trained on the original CCMatrix corpus (rows $6$ vs. $1$ and $12$ vs. $7$). Among the language-pairs considered, the largest improvements are reported for \textsc{it-oc} translation tasks (i.e., $+8.4/+6.7$), followed by \textsc{en-oc} (i.e., $+5.5/+4.4$). The magnitude of improvements might be explained by the relatedness of the two languages which facilitates editing with simpler operations (e.g., copying instead of translating). 

Our approach also brings significant improvements over the original data for distant language-pairs written in different scripts, despite being trained on more noisy data, as discussed above. For example, we see improvements $+2.0/+1.0$ for \textsc{en-be} and $+3.1/+3.7$ for \textsc{en-mr}. On the other hand, improvements on \textsc{en-sw} are smaller (i.e., $+0.5/+1.8$). This is expected given the high \textsc{bleu} scores that the original CCMatrix data yields. 

\paragraph{Comparison with Translation-based Baseline} Since Pool~B bitexts are typically
filtered out from the pool of \textsc{nmt} training instances, one reasonable way of incorporating them in \textsc{nmt} training is via treating them as monolingual samples. We experiment with a translation-based model that uses back-translation---the most popular approach to employ data augmentation for \textsc{nmt}. Comparing \textsc{nmt} models trained on CCMatrix augmented with back-translated Pool~B against our revised Pool~B version (i.e., rows $5$ vs. $6$ and $11$ vs. $12$) shows that editing outperforms the translation-based model for $7/10$ tasks, while it yields comparable results to it for the rest $3$. 

\begin{figure*}[!t]
    \centering
    \begin{subfigure}[t]{0.33\textwidth}
        \centering
        \includegraphics[height=1.2in]{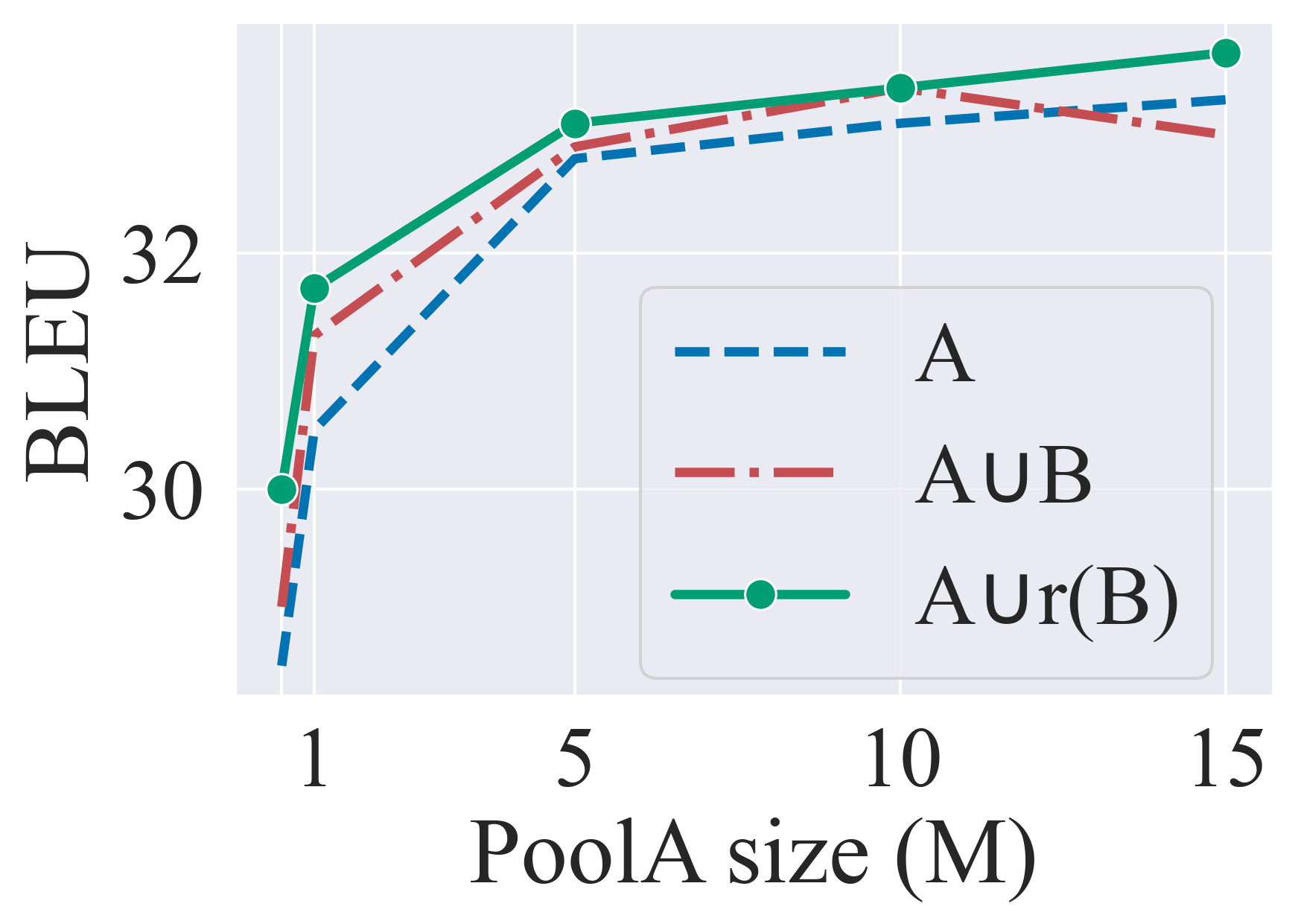}
        \caption{$|B| = |A|/2 $}\label{fig:scaling_a}
    \end{subfigure}%
    \hfill
    \begin{subfigure}[t]{0.33\textwidth}
        \centering
        \includegraphics[height=1.2in]{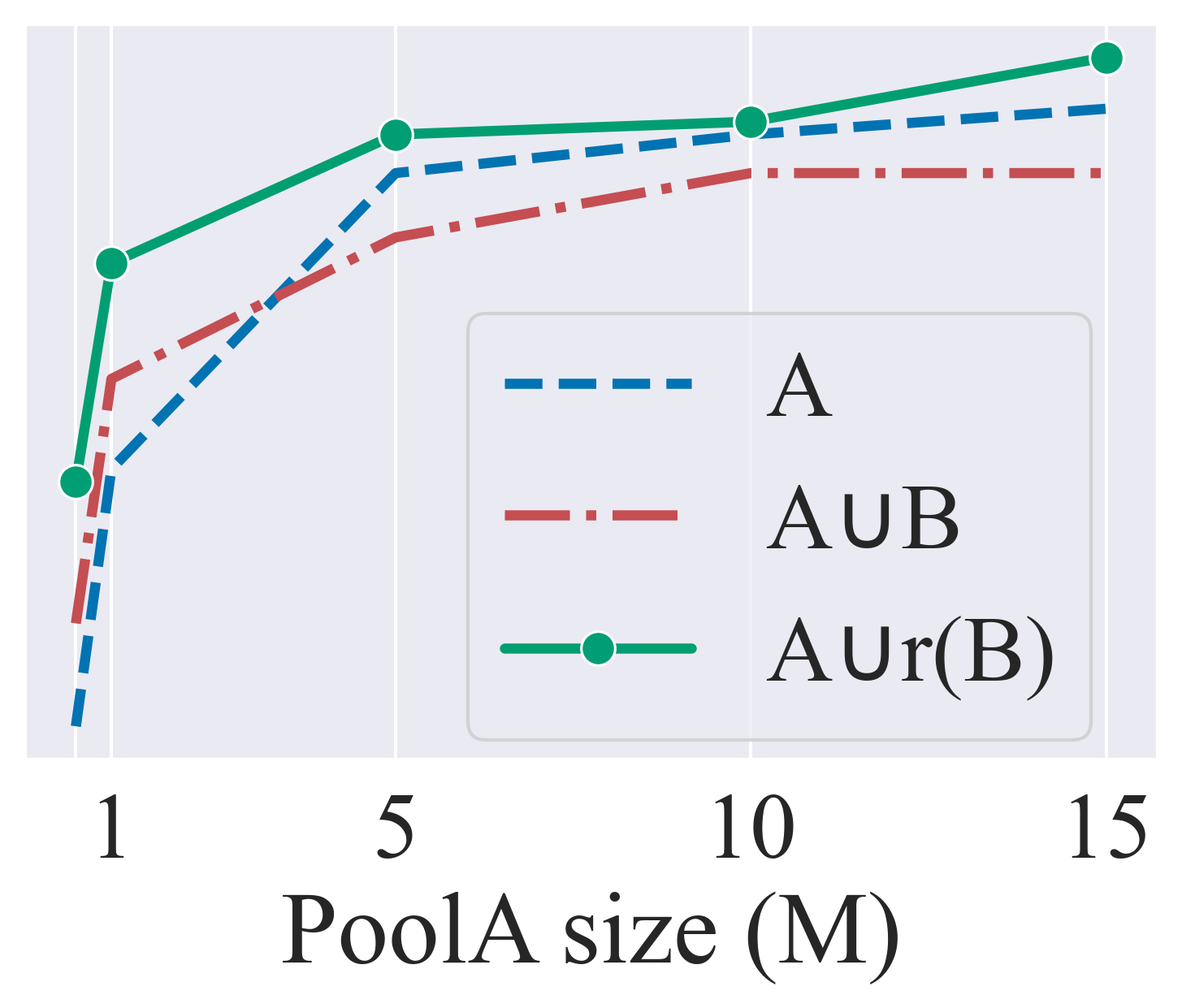}
        \caption{$|B| = |A| $}\label{fig:scaling_b}
    \end{subfigure}
    \hfill
    \begin{subfigure}[t]{0.33\textwidth}
        \centering
        \includegraphics[height=1.2in]{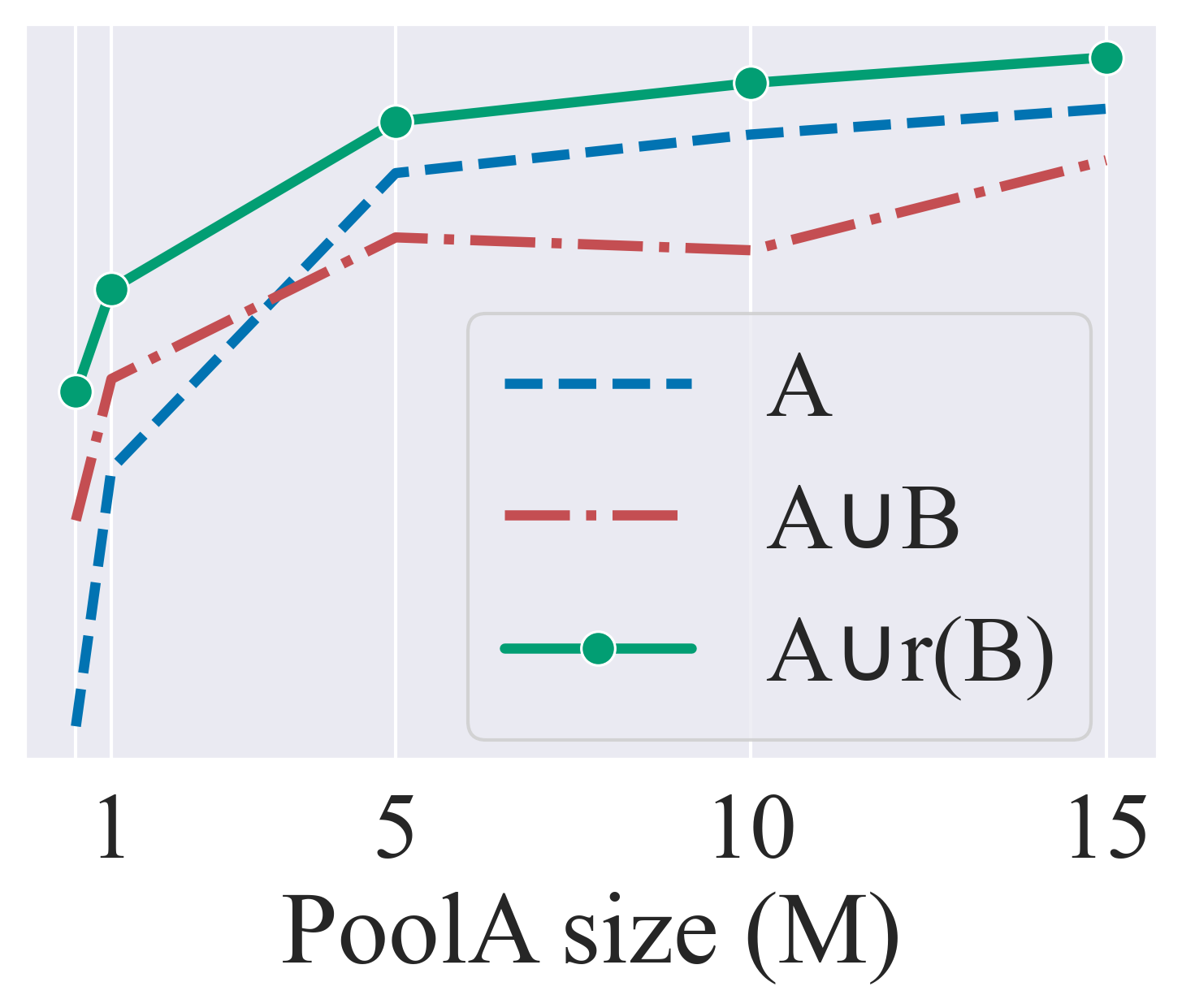}
        \caption{$|B| = 2|A| $}\label{fig:scaling_c}
    \end{subfigure}
\caption{Translation quality (i.e., \textsc{bleu}) of \textsc{en}$\rightarrow$\textsc{el} \textsc{nmt} models trained on  different amounts of Pool A and Pool B data (i.e., $|A|$ given by $x$-axis). Across settings, bitext refinement (i.e., $A\cup r(B)$) performs better or comparably to training on the original CCMatrix (i.e., $A\cup B$) or its filtered version (i.e., $A$).}%
    \label{fig:scaling_up}%
\end{figure*}

\paragraph{Editing Pool~A and Pool~B} Since the editing framework gives us the potential to generalize all types of operations that \textit{might} be needed to refine bitexts, it is also important that it does not perform \textit{overediting} (i.e., editing already good quality bitexts). For this reason, we also attempt to revise the entire CCMatrix corpus (i.e., $r(A\cup B)$), using our bitext refinement models (i.e., rows $4$ and $9$). To better understand the importance of performing conservative editing on good quality bitexts, we also compare against the translation-based baseline (i.e, $b(A\cup B)$ in  rows $3$ and $9$). First, we observe that our approach yields consistently significant improvements over CCMatrix with the exception of \textsc{en}$\rightarrow$\textsc{sw} where it performs comparably to it. Second, for most tasks the improvements are comparable to those reported when revising only Pool~B, while it is consistently better than the translation-based approach. It, overall, provides a universal method that works well in every case. 
%
%
\section{Analysis}
We now turn into analysis with a focus on understanding the broader space where \textsc{BitextEdit} can be applied. We experiment with scaling-up bitext refinement to higher-resource settings in \S\ref{sec:scaling_up}, we perform qualitative analysis on the edited bitexts in \S\ref{sec:qualitative}, and quantitative analysis on the types and intensity of edits in different corpora in \S\ref{sec:quantitative}.

\subsection{Scaling-up \textsc{BitextEdit}}\label{sec:scaling_up}

\begin{figure}[!t]
    \centering
    \includegraphics[scale=0.3]{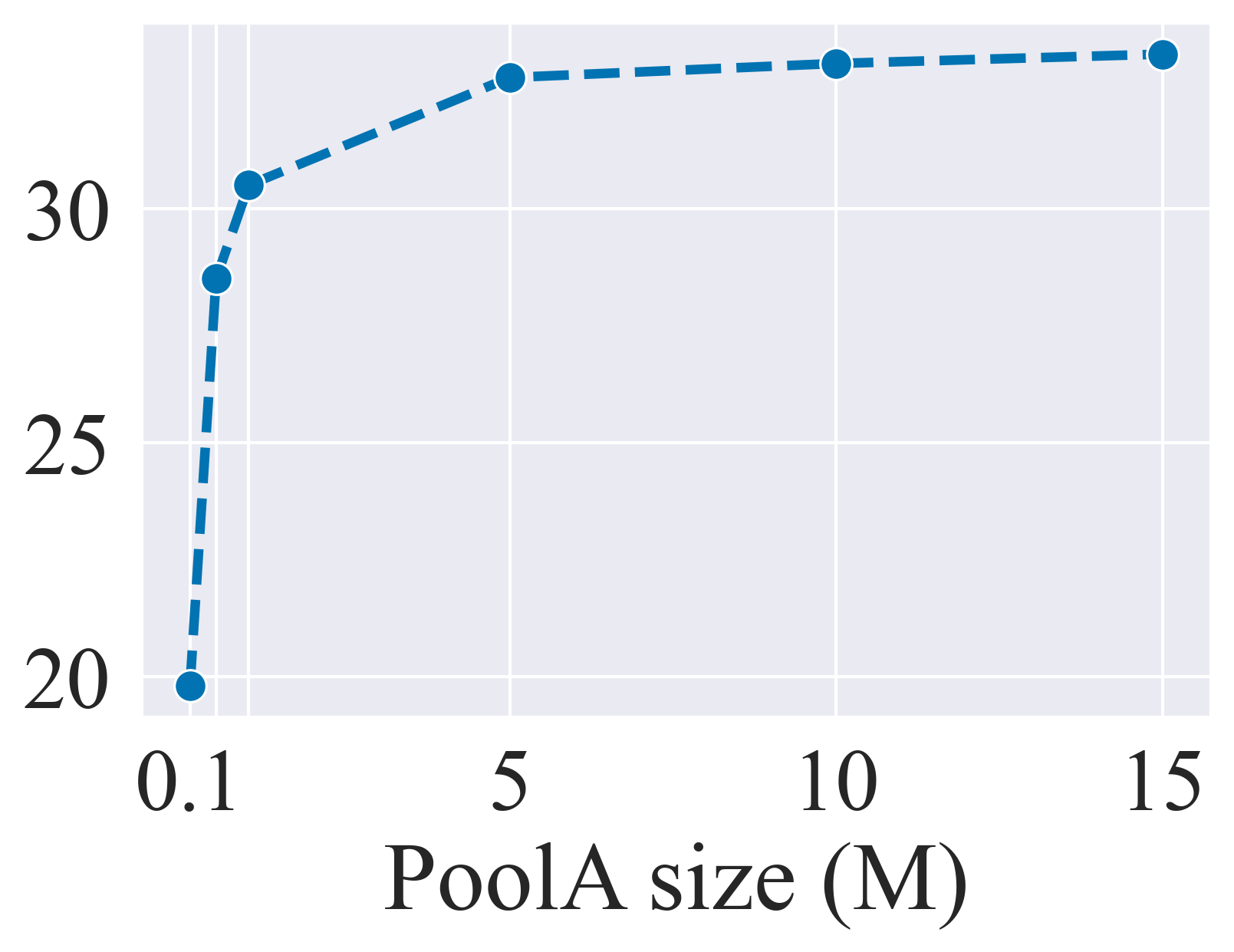}
    \caption{\textsc{bleu} for \textsc{en}$\rightarrow$\textsc{el} \textsc{nmt} trained on varying size of CCMatrix data (Pool A).}\vspace{-1em}
    \label{fig:bleu_curve_a}
\end{figure}

First, we examine how models trained only on good quality data (Figure~\ref{fig:bleu_curve_a}) behave as we vary their quantity. 
We experiment with English-Greek \textsc{en-el} CCMatrix bitexts and simulate various resource settings via downsampling. 
In \textit{low-resource} settings (i.e., $|A|<1$M), translation quality exhibits rapid improvements, with an increase from $100$K to $500$K training samples boosting \textsc{bleu}, by approximately $10$ points. In \textit{medium-resource} scenarios (i.e., $1<$M$|A|<5$M), a proportional increase in the quantity of good quality bitexts yields smaller---yet, significant---translation quality improvements (i.e., moving from $1$M to $5$M bitexts yields $+2$ \textsc{bleu}). Finally, 
in \textit{high-resource} settings (i.e, $|A|>5$), translation quality reaches a saturation point, with \textsc{bleu} increases being small and insignificant (i.e.,~$\sim +0.2$) as we move from $10$M to $15$M training samples. 

%
Second, we present a controlled analysis experiment on how bitext refinement impacts the translation quality of \textsc{nmt} systems under different resource settings (Figure~\ref{fig:scaling_up}). Starting from a high resource language-pair in CCMatrix (here, \textsc{en-el}) we sample good and poor quality bitexts (i.e, $A$ and $B$, respectively) representing low- to high- data scenarios (e.g, $500$K up to $15$M sentence-pairs). Then, we train \textsc{en}$\rightarrow$\textsc{el} \textsc{nmt} systems on $A\cup B$ while varying their distribution to represent three settings: (a) good quality bitexts overwhelm the training data (i.e, $|B| = |A|/2$), (b)  good and poor quality bitext are equally represented (i.e, $|B| = |A|$), and (c) poor quality bitexts overwhelm the training data (i.e, $|B| = 2|A|/$). We include more details on experimental settings in Appendix~\ref{sec:analysis_details}.

Across distribution conditions, adding imperfect translations (i.e., $B$) to the original good quality data yields improvements for low-to-medium resource settings (i.e, $|A|<5$). This results complement the earlier observations of \S\ref{sec:results} that question the appropriateness of a filtering framework in settings where data is scarce. On the other hand, when moving to high resource scenarios, the additional signal that results from imperfect references can have either insignificant (i.e., Figure~\ref{fig:scaling_a}) or negative impact (i.e., Figures~\ref{fig:scaling_b} and \ref{fig:scaling_c}) on translation quality. The latter depends on whether the good quality data is underrepresented in the training samples. 
\begin{table*}[!t]
    \centering
    \scalebox{0.67}{
    \begin{tabular}{l@{\hskip 0.3in}l@{\hskip 0.7in}l}
    %
    \\\toprule[2pt]
    \addlinespace[0.5em]
    
    %
    \editedside & \textbf{\textsc{[en]}} \small{\textsc{ccmatrix}} & \colorbox{red!10}{\hz After that time} the \colorbox{red!10}{\hz whole} group would \colorbox{red!10}{\hz talk} for 5 minutes. \\
    & \textbf{\textsc{[el]}} \small{\textsc{ccmatrix}} & 
    \selectlanguage{greek}
    Αργότερα, η ομάδα μελέτης ζήτησε από όλους να διαλογιστούν για πέντε λεπτά. \\
    & \hspace{0.5em} $\lfloor$ \hspace{0.3em} \small{\textsc{gloss}}  &
    {\color{gray!80}
    \hspace{0.2em}
    \textit{
    Later, the study group asked everyone to meditate for 5 minutes.}}\\
    & \textbf{\textsc{[en]}} \small{\textsc{BitextEdit}} & 
    \hspace{0.1em}
    Later, the study group asked everyone to meditate for five minutes. \\
    \addlinespace[0.8em]
    \cmidrule{2-3}
    \addlinespace[0.8em]
    \editedside & \textbf{\textsc{[en]}} \small{\textsc{ccmatrix}} &
    \hspace{0.1em}
    We should, however, always be striving to live a sustainable and \colorbox{red!10}{\hz kind life.} \\
    & \textbf{\textsc{[el]}} \small{\textsc{ccmatrix}} & 
    \selectlanguage{greek}
     Πάντα πρέπει να παλεύουμε για δίκαιη και βιώσιμη ειρήνη. \\
     & \hspace{0.5em} $\lfloor$ \hspace{0.3em} \small{\textsc{gloss}}  &
    {\color{gray!80}
    \hspace{0.2em}
    \textit{ We must always fight for a just and lasting peace.
    }}\\        %
    & \textbf{\textsc{[en]}} \small{\textsc{BitextEdit}} & 
    \hspace{0.1em}
     We must always fight for just and sustainable peace. \\

    \addlinespace[0.8em]
    \cmidrule{2-3}
    \addlinespace[0.8em]

    & \textbf{\textsc{[en]}} \small{\textsc{ccmatrix}} & 
    ``The western influence came from film and television'', he later explained. \\
    \editedside & \textbf{\textsc{[en]}} \small{\textsc{ccmatrix}} &
    \selectlanguage{greek} 
    «Η \colorbox{red!10}{\hz λογοκρισία εντείνεται όλο και περισσότερο στον κινηματογράφο} και την τηλεόραση», εξήγησε ο ίδιος. \\
     & \hspace{0.5em} $\lfloor$ \hspace{0.3em} \small{\textsc{gloss}}  &
    {\color{gray!80}
    \hspace{0.2em}
    \textit{
    ``Censorship is intensifying in cinema and television'', he explained.
    }}\\
    & \textbf{\textsc{[el]}} \small{\textsc{BitextEdit}} & 
    \selectlanguage{greek}
     «Η δυτική επιρροή ήρθε από την ταινία και την τηλεόραση» , εξήγησε αργότερα. \\
    & \hspace{0.5em} $\lfloor$ \hspace{0.3em} \small{\textsc{gloss}}  &
    {\color{gray!80}
    \hspace{0.2em}
    \textit{
    ``The western influence came from form and television'', as their later explained.
    }}\\
    \addlinespace[0.8em]
    \cmidrule{2-3}
    \addlinespace[0.8em]
    & \textbf{\textsc{[en]}} \small{\textsc{ccmatrix}} &
    \hspace{0.1em}
    I could work with a hospital specialist as a clinical assistant (as I have done). \\
    \editedside & \textbf{\textsc{[el]}} \small{\textsc{ccmatrix}} &
    \selectlanguage{greek}
    \colorbox{red!10}{\hz Δούλευε ώς βοηθός ερευνητή παράλληλα με το διδακτορικό (όπως και εγώ)} \\
     & \hspace{0.5em} $\lfloor$ \hspace{0.3em} \small{\textsc{gloss}}  &
    {\color{gray!80}
    \hspace{0.2em}
    \textit{ They were working as an assistant researcher in parallel with their doctorate (as I have done).
    }}\\        %
    & \textbf{\textsc{[el]}} \small{\textsc{BitextEdit}} & 
    \selectlanguage{greek}
     Θα μπορούσα να δουλέψω με έναν ειδικό στο νοσοκομείο ως κλινικός βοηθός (όπως έχω κάνει).\\
     & \hspace{0.5em} $\lfloor$ \hspace{0.3em} \small{\textsc{gloss}}  &
    {\color{gray!80}
    \hspace{0.2em}
    \textit{ I could work with a hospital specialist as a clinical assistant (as I have done).
    }}   %
    \\\toprule[2pt]
    \end{tabular}}
    \caption{Examples of CCMatrix bitexts along with refined sides generated by \textsc{BitextEdit}. \editedside \ \  denotes the side ([\textsc{el}] or [\textsc{en}]) that the model edits, while highlighted segments indicate the meaning mismatches in the original CCMatrix sentence that gets edited. Greek sentences are glossed to help understanding their meaning.}
    \label{tab:model_outputs_main}
\end{table*}

Third, starting from good quality bitexts of varying sizes, we train separate bitext refinement models and edit the corresponding poor quality samples (i.e., $r(.)$) defined earlier. Across the board, \textsc{nmt} models that are trained on $A \cup r(B)$ yield the best translation quality results compared to both filtering and training on original CCMatrix. However, we observe that the magnitude of the improvements depends on the data settings. Concretely, bitext refinement yields significant improvements on low-to-medium resource settings (i.e., $\sim+2$ \textsc{blue} points).  On the other hand, in high resource scenarios bitext refinement helps mitigate the negative impact of overwhelming poor quality instances and performs comparably to filtering. The latter suggests that our refinement strategy \textit{improves bitexts quality across low- to high- resource settings}.

\subsection{Qualitative analysis}\label{sec:qualitative}

We conduct a qualitative study to confirm that \textsc{BitextEdit} improves the quality of CCMatrix. We include details on the annotation in Appendix~\ref{sec:manual_annotation_details}. 
One of the authors manually evaluates a random sample of $200$ \textsc{en-el} sentence-pairs where we compare the original bitexts against the refined ones. Here, we present results on bitext refinement models that use $0.5$M PoolA samples. Manual inspection on refined outputs of models trained on larger pools showed similar performance.
\begin{figure}[!t]
    \centering
    \includegraphics[scale=0.25]{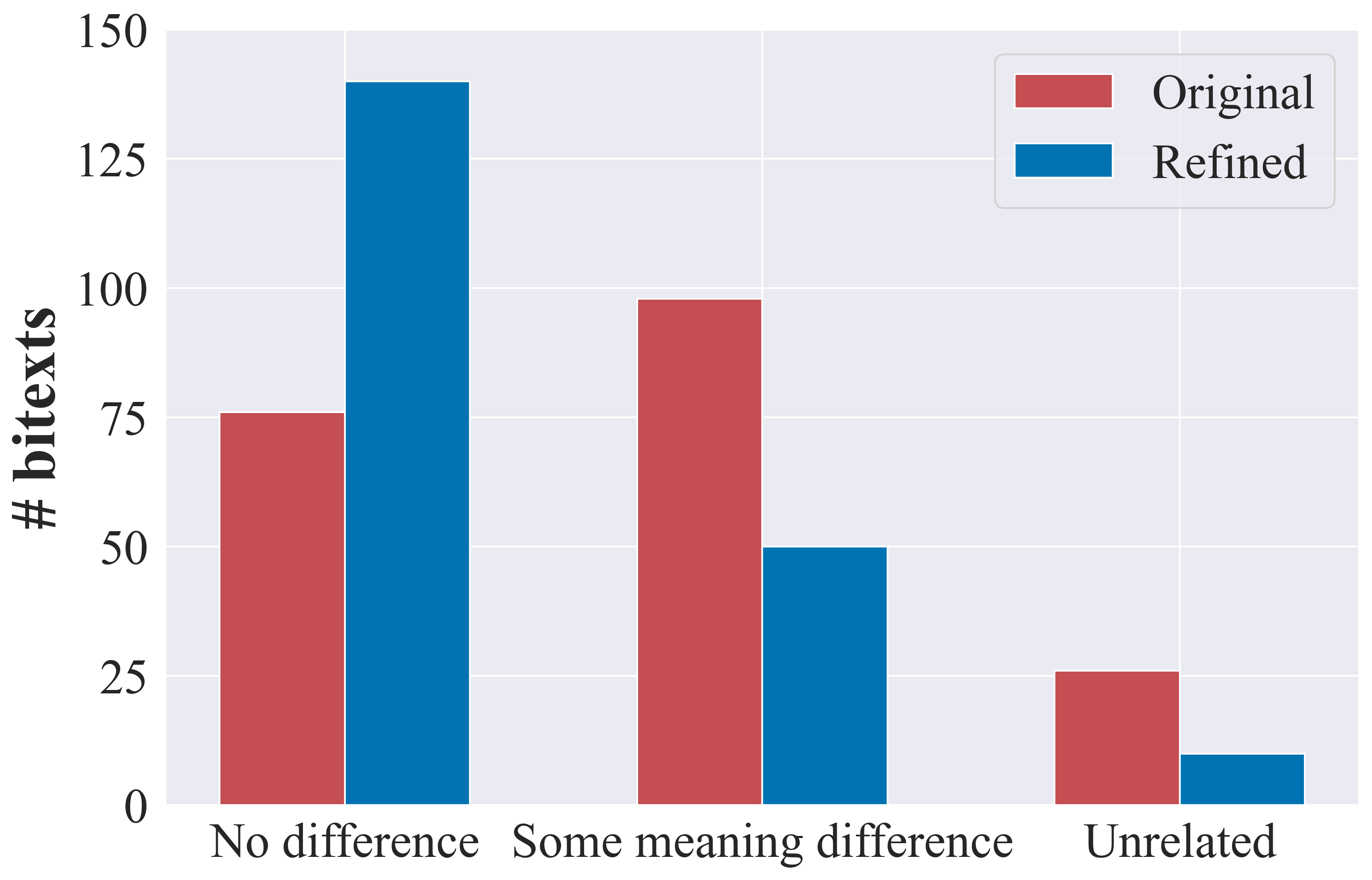}
    \caption{Number of bitexts manually rated as perfect translations (i.e., No difference), partial translations (i.e., some meaning difference), and wrong translations (i.e., unrelated) for a random sample of original vs. refined CCMatrix \textsc{en-el} data.}
    \label{fig:qual}
\end{figure}
%
As shown in Figure~\ref{fig:qual}, our models performs edits that refine meaning mismatches found in the original CCMatrix data. While only $~\sim 38\%$ of the original samples contain parallel texts that are perfect translations of each other, the revised sample contains $~\sim 70\%$ perfect translations. 
Finally---apart from evaluating meaning differences---we also rate fluency of the edited translations. We find that our model does not suffer from major fluency issues with $84.5\%$ of their outputs rated as \textit{flawless} and $15.5\%$ as \textit{good}. Table~\ref{tab:model_outputs_main} presents example outputs of our \textsc{BitextEdit} approach for English-Greek. More examples can be found in Appendix~\ref{sec:model_outputs}.
\begin{table*}[!t]
    \centering
    \scalebox{0.9}{
    \begin{tabular}{lr@{\hskip 0.3in}rrrr@{\hskip 0.4in}rrrr}
    & & \textbf{C} & \textbf{S} & \textbf{D} & \textbf{I} & \textbf{C} & \textbf{S} & \textbf{D} & \textbf{I} \\
    \rowcolor{gray!10}
    \textsc{\textbf{corpus}} & \textbf{\textsc{edited sent.}} & \multicolumn{4}{c}{\textbf{\textsc{all ($\%$)}}} & \multicolumn{4}{c}{\textbf{\textsc{all }}$\setminus$ \textsc{\textbf{copies}} ($\%$)} \\
    \centering
    Tatoeba       & $29.80\%$ & $97.47$ & $1.88$ & 	$0.29$ &  $0.34$ & $86.38$ & $10.16$ & $1.56$ & $1.88$  \\
    OpenSubtitles & $65.63\%$ & $90.46$ & $5.53$ &	$1.27$ &  $2.73$ & $74.51$ & $14.79$ & $3.39$ & $7.29$ \\
    ParaCrawl     & $88.11\%$ & $96.30$ & $2.25$ &	$0.39$ &  $1.04$ & $85.42$ & $8.89$ &	$1.55$ & $4.12$ \\
    \end{tabular}}
    \caption{\textsc{ter} statistics for bitext refinement of random samples of \textsc{en-el} \textsc{opus} bitexts. Second column gives the $\%$ of bitexts that get at least one edit operation; the last two columns present the percentage of correct~(\textsc{\textbf{c}}), substituted~(\textsc{\textbf{s}}), deleted~(\textsc{\textbf{d}}), and inserted~(\textsc{\textbf{i}}) tokens for all the bitexts (i.e., \textbf{\textsc{all}}) and the subset of bitexts that receive revisions compared to the original (i.e., \textbf{\textsc{all}}$\setminus$ \textsc{\textbf{copies}}). }
    \label{tab:ter_labels}
\end{table*}

\subsection{Quantitative analysis}\label{sec:quantitative}
\paragraph{Percentage of edited bitexts} Table~\ref{tab:copies} presents coarse statistics on the percentage of refined bitexts that exhibit at least one edit compared to the original ones. 
First, we observe that the percentage of edited bitexts varies across the languages-pairs studied. This reflects the varying quality of PoolB samples in different languages and also
connects to the varying magnitude of improvements we show in Table~\ref{tab:main_results}. The biggest improvements are given for \textsc{it-oc}, where $\sim 76\%$ of the bitexts are edited by our refinement models. On the other hand, the smallest improvements are found for \textsc{en-sw}, with only $\sim 36\%$ of its bitext being revised, probably due to the already good quality of the initial CCMatrix sentence pairs. 

\paragraph{Editing \textsc{en-el} \textsc{opus} corpora}
Broadly speaking, a good bitext refinement model should be able to rewrite bitext in a way that improves potential errors in the original references. At the same time though, it should avoid over-editing (i.e., avoid editing an already good translation-pair). We perform a quantitative analysis on  \textsc{en-el} corpora from \textsc{opus} that vary in their quality and extract Translation Error Rate (\textsc{ter}) label \cite{snover-etal-2006-study} token-level statistics to study both the \textit{frequency} and the \textit{types} of edits that our bitext refinement models perform. Table~\ref{tab:ter_labels} presents  results on random samples ($\sim$ 100K) of three popular corpora: 
(a) the Tatoeba corpus~\cite{tiedemann-2020-tatoeba} consisting of human translations, 
(b) the OpenSubtitles corpus~\cite{lison-tiedemann-2016-opensubtitles2016} consisting of sentence-aligned subtitles of movie series\footnote{\url{http://www.opensubtitles.org/}, \url{https://opus.nlpl.eu/OpenSubtitles-v2018.php}}, 
and (c) the ParaCrawl corpus~\cite{espla-etal-2019-paracrawl} consisting of automatically crawled translations from translations of European Parliament Proceedings.
\begin{table}[!t]
    \centering
    \scalebox{0.9}{
    \begin{tabular}{lccc}
    \rowcolor{gray!10}
    \textbf{\textsc{pair}} & \textbf{\textsc{src}} & \textbf{\textsc{tgt}} & \textbf{\textsc{both}} \\
    \addlinespace[0.2em]
    \textsc{en-oc} &  $34.06\%$ &  $66.58\%$ & $67.48\%$ \\
    \textsc{it-oc} &  $34.76\%$ &  $41.11\%$ & $75.78\%$ \\ 
    \textsc{en-mr} &  $58.35\%$ &  $19.90\%$ & $68.07\%$ \\
    \textsc{be-en} &  $21.01\%$ &  $28.06\%$ & $49.06\%$ \\
    \textsc{en-sw} &  $14.52\%$ &  $21.05\%$ & $35.57\%$ \\
    \end{tabular}}
    \caption{Percentage of sentences with at least one edit operation compared to the original for: source-side (\textsc{src}), target-side (\textsc{tgt}), and both sides (\textsc{both}).}
    \label{tab:copies}
\end{table}

As expected, our model performs minimal editing on the high-quality \textit{Tatoeba} bitexts. Concretely, only $\sim 30\%$ of it gets revised, while as suggested by the token-level \textsc{ter} statistics even the revised sentence-pairs mostly consist of substituted tokens. Further manual inspection reveals that most of those tokens depict subtle spelling differences between Greek words. On the other hand, when editing the samples of automatically extracted bitexts our refinement model performs more frequent edits: it revises $\sim65\%$ of OpenSubtitles and $\sim 88\%$ of ParaCrawl bitexts. Interestingly, although a greater amount of ParaCrawl texts get revised compared to OpenSubtitles, edits on the latter are more aggressive as it consists of at least $10\%$ fewer correct (i.e., \textsc{\textbf{c}}) tokens than the former. A break down on the types of operations further reveals that editing OpenSubtitles requires more deletion (i.e., \textsc{\textbf{d}}) and insertion  (i.e., \textsc{\textbf{i}})
operations compared to the other two. This observation connects
to prior efforts on auditing OpenSubtitles that found sentence segmentation errors (i.e, added extra leading/trailing words in one side) to be a frequent type error for this corpus~\cite{vyas-etal-2018-identifying}.
%
%
\section{Related Work}

\paragraph{Automatic Post-Editing} \textsc{ape} aims at automatically correcting the output of a black-box \textsc{mt} system. 
Recent approaches on \textsc{ape}~\cite{chatterjee-etal-2019-findings, chatterjee-etal-2020-findings} 
fine-tune pre-trained multilingual models models~\cite{lopes-etal-2019-unbabels} or translation models~\cite{yang-etal-2020-hw} on a combination of gold-standard \textsc{ape} data and artificially augmented candidates resulting from external translations. 
\textsc{BitextEdit} aims instead, at editing imperfect translations representing human generated texts in two languages, without assuming access to gold-standard training data.

\paragraph{Low-resource \textsc{mt}} \citet{haddow2021survey} structure the diverse set of approaches to low-resource \textsc{mt} to (a) 
efforts for increasing the amounts of available bitexts (i.e., \textit{data collection}; ~\citet{schwenk-etal-2021-wikimatrix,schwenk-etal-2021-ccmatrix}), 
(b) methods that explore how other types of data
can be incorporated into \textsc{mt} (i.e., \textit{data exploitation};~\citet{baziotis-etal-2020-language,zoph-etal-2016-transfer,garcia2018}), 
and (c) advances in modeling (i.e., \textit{model choices};~\citet{NIPS2017_3f5ee243}). \textsc{BitextEdit} is an alernative \textit{data exploitation} approach that does not require further bilingual data or other sources of supervision.

\paragraph{Synthetic Bitext} Generating synthetic bitext has mainly been studied as a means of data augmentation for \textsc{nmt} through forward translation~\cite{zhang-zong-2016-exploiting}, backtranslation~\cite{sennrich-etal-2016-improving, marie-etal-2020-tagged, hoang-etal-2018-iterative}, or round-trip translation~\cite{Ahmadnia2019AugmentingNM}  of monolingual resources. Moreover, recent line of works use the predictions of forward and backward translation models to induce the creation of new versions of the parallel data: \citet{10.5555/3495724.3496564} diversify the parallel data via translating both sides using multiple models and then merge them with the original to train a ﬁnal NMT model; \citet{jiao-etal-2020-data} employ a similar approach to rejuvenate the most inactive examples that contributes less to the model performance; \citet{kim-rush-2016-sequence} propose to train a student model of smaller capacity on sequence-level interpolated data generated by a teacher model of higher capacity.
Using synthetic translations to augment or revise real bitexts assumes access to \textsc{nmt} systems of sufficient quality. Recent works propose methods to automatically revise noisy synthetic bitexts~\cite{Cheng2020ARAT,wei-etal-2020-iterative}. 
By contrast, our work accounts for imperfect references in \textit{real} bitext and is tailored to low-resource settings where \textsc{
nmt} quality is too low to provide reliable candidate translations. 

\paragraph{Retrieve \& Edit Approaches} Retrieve and edit approaches have been integrated at inference time for several tasks, such as \textsc{nmt}~\cite{Gu_Wang_Cho_Li_2018,zhang-etal-2018-guiding, cao-xiong-2018-encoding,hossain-etal-2020-simple}, \textsc{ape}~\cite{hokamp-2017-ensembling}, dialogue generation~\cite{weston-etal-2018-retrieve}, among others.

%
%
\section{Conclusion}\label{sec:conclusion}

We introduce an alternative approach for bitext quality improvement that we show is better suited for low-resource language pairs. Instead of filtering out imperfect translation references that result from automatic bitext mining, we instead edit them with the goal of improving their quality. Our editing models are trained using only synthetic supervision, which can be gathered at scale for any language pair that support bitext mining. Extensive quantitative analysis suggests that our approach successfully improves bitext quality for a variety of language-pairs and different resource conditions. Furthermore, extrinsic experiments on $10$ low-resource \textsc{nmt} tasks suggest that bitext refinement constitutes a successful approach to improving \textsc{nmt} translation quality in low data regimes. Those findings
highlight the importance of the good \textit{quality} bitexts in scenarios where large \textit{quantities} cannot be guaranteed
and motivate future research on improving low-resource \textsc{nmt} further.

\bibliography{anthology,custom}

\begin{thebibliography}{65}
\expandafter\ifx\csname natexlab\endcsname\relax\def\natexlab#1{#1}\fi

\bibitem[{Ahmadnia and Dorr(2019)}]{Ahmadnia2019AugmentingNM}
Benyamin Ahmadnia and B.~Dorr. 2019.
\newblock Augmenting neural machine translation through round-trip training
  approach.
\newblock \emph{Open Computer Science}, 9:268 -- 278.

\bibitem[{Artetxe and Schwenk(2019)}]{artetxe-schwenk-2019-massively}
Mikel Artetxe and Holger Schwenk. 2019.
\newblock \href {https://doi.org/10.1162/tacl_a_00288} {Massively multilingual
  sentence embeddings for zero-shot cross-lingual transfer and beyond}.
\newblock \emph{Transactions of the Association for Computational Linguistics},
  7:597--610.

\bibitem[{Ash et~al.(2018)Ash, Francis, and
  Williams}]{ash-etal-2018-speechmatics}
Tom Ash, Remi Francis, and Will Williams. 2018.
\newblock \href {https://doi.org/10.18653/v1/W18-6472} {The speechmatics
  parallel corpus filtering system for {WMT}18}.
\newblock In \emph{Proceedings of the Third Conference on Machine Translation:
  Shared Task Papers}, pages 853--859, Belgium, Brussels. Association for
  Computational Linguistics.

\bibitem[{Ba{\~n}{\'o}n et~al.(2020)Ba{\~n}{\'o}n, Chen, Haddow, Heafield,
  Hoang, Espl{\`a}-Gomis, Forcada, Kamran, Kirefu, Koehn, Ortiz~Rojas,
  Pla~Sempere, Ram{\'\i}rez-S{\'a}nchez, Sarr{\'\i}as, Strelec, Thompson,
  Waites, Wiggins, and Zaragoza}]{banon-etal-2020-paracrawl}
Marta Ba{\~n}{\'o}n, Pinzhen Chen, Barry Haddow, Kenneth Heafield, Hieu Hoang,
  Miquel Espl{\`a}-Gomis, Mikel~L. Forcada, Amir Kamran, Faheem Kirefu, Philipp
  Koehn, Sergio Ortiz~Rojas, Leopoldo Pla~Sempere, Gema
  Ram{\'\i}rez-S{\'a}nchez, Elsa Sarr{\'\i}as, Marek Strelec, Brian Thompson,
  William Waites, Dion Wiggins, and Jaume Zaragoza. 2020.
\newblock \href {https://doi.org/10.18653/v1/2020.acl-main.417} {{P}ara{C}rawl:
  Web-scale acquisition of parallel corpora}.
\newblock In \emph{Proceedings of the 58th Annual Meeting of the Association
  for Computational Linguistics}, pages 4555--4567, Online. Association for
  Computational Linguistics.

\bibitem[{Baziotis et~al.(2020)Baziotis, Haddow, and
  Birch}]{baziotis-etal-2020-language}
Christos Baziotis, Barry Haddow, and Alexandra Birch. 2020.
\newblock \href {https://doi.org/10.18653/v1/2020.emnlp-main.615} {Language
  model prior for low-resource neural machine translation}.
\newblock In \emph{Proceedings of the Conference on Empirical Methods in
  Natural Language Processing (EMNLP)}, pages 7622--7634, Online. Association
  for Computational Linguistics.

\bibitem[{Briakou and Carpuat(2020)}]{briakou-carpuat-2020-detecting}
Eleftheria Briakou and Marine Carpuat. 2020.
\newblock \href {https://doi.org/10.18653/v1/2020.emnlp-main.121} {{D}etecting
  {F}ine-{G}rained {C}ross-{L}ingual {S}emantic {D}ivergences without
  {S}upervision by {L}earning to {R}ank}.
\newblock In \emph{Proceedings of the 2020 Conference on Empirical Methods in
  Natural Language Processing (EMNLP)}, pages 1563--1580, Online. Association
  for Computational Linguistics.

\bibitem[{Briakou and Carpuat(2021)}]{briakou-carpuat-2021-beyond}
Eleftheria Briakou and Marine Carpuat. 2021.
\newblock \href {https://doi.org/10.18653/v1/2021.acl-long.562} {Beyond noise:
  Mitigating the impact of fine-grained semantic divergences on neural machine
  translation}.
\newblock In \emph{Proceedings of the 59th Annual Meeting of the Association
  for Computational Linguistics and the 11th International Joint Conference on
  Natural Language Processing (Volume 1: Long Papers)}, pages 7236--7249,
  Online. Association for Computational Linguistics.

\bibitem[{Briakou and Carpuat(2022)}]{briakou2022}
Eleftheria Briakou and Marine Carpuat. 2022.
\newblock \href {https://arxiv.org/pdf/2203.07643.pdf} {Can synthetic
  translations improve bitext quality?}
\newblock In \emph{Proceedings of the 60th Annual Meeting of the Association
  for Computational Linguistics \textit{(to appear)}}, Online. Association for
  Computational Linguistics.

\bibitem[{Cao and Xiong(2018)}]{cao-xiong-2018-encoding}
Qian Cao and Deyi Xiong. 2018.
\newblock \href {https://doi.org/10.18653/v1/D18-1340} {Encoding gated
  translation memory into neural machine translation}.
\newblock In \emph{Proceedings of the 2018 Conference on Empirical Methods in
  Natural Language Processing}, pages 3042--3047, Brussels, Belgium.
  Association for Computational Linguistics.

\bibitem[{Chatterjee et~al.(2019)Chatterjee, Federmann, Negri, and
  Turchi}]{chatterjee-etal-2019-findings}
Rajen Chatterjee, Christian Federmann, Matteo Negri, and Marco Turchi. 2019.
\newblock \href {https://doi.org/10.18653/v1/W19-5402} {Findings of the {WMT}
  2019 shared task on automatic post-editing}.
\newblock In \emph{Proceedings of the Fourth Conference on Machine Translation
  (Volume 3: Shared Task Papers, Day 2)}, pages 11--28, Florence, Italy.
  Association for Computational Linguistics.

\bibitem[{Chatterjee et~al.(2020)Chatterjee, Freitag, Negri, and
  Turchi}]{chatterjee-etal-2020-findings}
Rajen Chatterjee, Markus Freitag, Matteo Negri, and Marco Turchi. 2020.
\newblock \href {https://aclanthology.org/2020.wmt-1.75} {Findings of the {WMT}
  2020 shared task on automatic post-editing}.
\newblock In \emph{Proceedings of the Fifth Conference on Machine Translation},
  pages 646--659, Online. Association for Computational Linguistics.

\bibitem[{Chaudhary et~al.(2019)Chaudhary, Tang, Guzm{\'a}n, Schwenk, and
  Koehn}]{chaudhary-etal-2019-low}
Vishrav Chaudhary, Yuqing Tang, Francisco Guzm{\'a}n, Holger Schwenk, and
  Philipp Koehn. 2019.
\newblock \href {https://doi.org/10.18653/v1/W19-5435} {Low-resource corpus
  filtering using multilingual sentence embeddings}.
\newblock In \emph{Proceedings of the Fourth Conference on Machine Translation
  (Volume 3: Shared Task Papers, Day 2)}, pages 261--266, Florence, Italy.
  Association for Computational Linguistics.

\bibitem[{Cheng et~al.(2020)Cheng, Kuang, Weng, Yu, Zhu, and
  Luo}]{Cheng2020ARAT}
Shanbo Cheng, Shaohui Kuang, Rongxiang Weng, Heng Yu, Changfeng Zhu, and Weihua
  Luo. 2020.
\newblock Ar: Auto-repair the synthetic data for neural machine translation.
\newblock \emph{ArXiv}, abs/2004.02196.

\bibitem[{Conneau and Lample(2019)}]{NEURIPS2019_c04c19c2}
Alexis Conneau and Guillaume Lample. 2019.
\newblock \href
  {https://proceedings.neurips.cc/paper/2019/file/c04c19c2c2474dbf5f7ac4372c5b9af1-Paper.pdf}
  {Cross-lingual language model pretraining}.
\newblock In \emph{Advances in Neural Information Processing Systems},
  volume~32. Curran Associates, Inc.

\bibitem[{El-Kishky et~al.(2020)El-Kishky, Chaudhary, Guzm{\'a}n, and
  Koehn}]{el-kishky-etal-2020-ccaligned}
Ahmed El-Kishky, Vishrav Chaudhary, Francisco Guzm{\'a}n, and Philipp Koehn.
  2020.
\newblock \href {https://doi.org/10.18653/v1/2020.emnlp-main.480} {{CCA}ligned:
  A massive collection of cross-lingual web-document pairs}.
\newblock In \emph{Proceedings of the 2020 Conference on Empirical Methods in
  Natural Language Processing (EMNLP)}, pages 5960--5969, Online. Association
  for Computational Linguistics.

\bibitem[{Espl{\`a} et~al.(2019)Espl{\`a}, Forcada, Ram{\'\i}rez-S{\'a}nchez,
  and Hoang}]{espla-etal-2019-paracrawl}
Miquel Espl{\`a}, Mikel Forcada, Gema Ram{\'\i}rez-S{\'a}nchez, and Hieu Hoang.
  2019.
\newblock \href {https://aclanthology.org/W19-6721} {{P}ara{C}rawl: Web-scale
  parallel corpora for the languages of the {EU}}.
\newblock In \emph{Proceedings of Machine Translation Summit XVII: Translator,
  Project and User Tracks}, pages 118--119, Dublin, Ireland. European
  Association for Machine Translation.

\bibitem[{Garc{\'{\i}}a{-}Mart{\'{\i}}nez
  et~al.(2017)Garc{\'{\i}}a{-}Mart{\'{\i}}nez, Barrault, and
  Bougares}]{garcia2018}
Mercedes Garc{\'{\i}}a{-}Mart{\'{\i}}nez, Lo{\"{\i}}c Barrault, and Fethi
  Bougares. 2017.
\newblock \href {http://arxiv.org/abs/1712.01821} {Neural machine translation
  by generating multiple linguistic factors}.
\newblock \emph{CoRR}, abs/1712.01821.

\bibitem[{Gu et~al.(2018)Gu, Wang, Cho, and Li}]{Gu_Wang_Cho_Li_2018}
Jiatao Gu, Yong Wang, Kyunghyun Cho, and Victor~O.K. Li. 2018.
\newblock \href {https://ojs.aaai.org/index.php/AAAI/article/view/12013}
  {Search engine guided neural machine translation}.
\newblock \emph{Proceedings of the AAAI Conference on Artificial Intelligence},
  32(1).

\bibitem[{Guzm{\'a}n et~al.(2019)Guzm{\'a}n, Chen, Ott, Pino, Lample, Koehn,
  Chaudhary, and Ranzato}]{guzman-etal-2019-flores}
Francisco Guzm{\'a}n, Peng-Jen Chen, Myle Ott, Juan Pino, Guillaume Lample,
  Philipp Koehn, Vishrav Chaudhary, and Marc{'}Aurelio Ranzato. 2019.
\newblock \href {https://doi.org/10.18653/v1/D19-1632} {The {FLORES} evaluation
  datasets for low-resource machine translation: {N}epali{--}{E}nglish and
  {S}inhala{--}{E}nglish}.
\newblock In \emph{Proceedings of the 2019 Conference on Empirical Methods in
  Natural Language Processing and the 9th International Joint Conference on
  Natural Language Processing (EMNLP-IJCNLP)}, pages 6098--6111, Hong Kong,
  China. Association for Computational Linguistics.

\bibitem[{Haddow et~al.(2021)Haddow, Bawden, Barone, Helcl, and
  Birch}]{haddow2021survey}
Barry Haddow, Rachel Bawden, Antonio Valerio~Miceli Barone, Jindřich Helcl,
  and Alexandra Birch. 2021.
\newblock \href {http://arxiv.org/abs/2109.00486} {Survey of low-resource
  machine translation}.

\bibitem[{Heilman et~al.(2014)Heilman, Cahill, Madnani, Lopez, Mulholland, and
  Tetreault}]{heilman-etal-2014-predicting}
Michael Heilman, Aoife Cahill, Nitin Madnani, Melissa Lopez, Matthew
  Mulholland, and Joel Tetreault. 2014.
\newblock \href {https://doi.org/10.3115/v1/P14-2029} {Predicting
  grammaticality on an ordinal scale}.
\newblock In \emph{Proceedings of the 52nd Annual Meeting of the Association
  for Computational Linguistics (Volume 2: Short Papers)}, pages 174--180,
  Baltimore, Maryland. Association for Computational Linguistics.

\bibitem[{Hoang et~al.(2018)Hoang, Koehn, Haffari, and
  Cohn}]{hoang-etal-2018-iterative}
Vu~Cong~Duy Hoang, Philipp Koehn, Gholamreza Haffari, and Trevor Cohn. 2018.
\newblock \href {https://doi.org/10.18653/v1/W18-2703} {Iterative
  back-translation for neural machine translation}.
\newblock In \emph{Proceedings of the 2nd Workshop on Neural Machine
  Translation and Generation}, pages 18--24, Melbourne, Australia. Association
  for Computational Linguistics.

\bibitem[{Hokamp(2017)}]{hokamp-2017-ensembling}
Chris Hokamp. 2017.
\newblock \href {https://doi.org/10.18653/v1/W17-4775} {Ensembling factored
  neural machine translation models for automatic post-editing and quality
  estimation}.
\newblock In \emph{Proceedings of the Second Conference on Machine
  Translation}, pages 647--654, Copenhagen, Denmark. Association for
  Computational Linguistics.

\bibitem[{Hossain et~al.(2020)Hossain, Ghazvininejad, and
  Zettlemoyer}]{hossain-etal-2020-simple}
Nabil Hossain, Marjan Ghazvininejad, and Luke Zettlemoyer. 2020.
\newblock \href {https://doi.org/10.18653/v1/2020.acl-main.228} {Simple and
  effective retrieve-edit-rerank text generation}.
\newblock In \emph{Proceedings of the 58th Annual Meeting of the Association
  for Computational Linguistics}, pages 2532--2538, Online. Association for
  Computational Linguistics.

\bibitem[{Jiao et~al.(2020)Jiao, Wang, He, King, Lyu, and
  Tu}]{jiao-etal-2020-data}
Wenxiang Jiao, Xing Wang, Shilin He, Irwin King, Michael Lyu, and Zhaopeng Tu.
  2020.
\newblock \href {https://doi.org/10.18653/v1/2020.emnlp-main.176} {{D}ata
  {R}ejuvenation: {E}xploiting {I}nactive {T}raining {E}xamples for {N}eural
  {M}achine {T}ranslation}.
\newblock In \emph{Proceedings of the 2020 Conference on Empirical Methods in
  Natural Language Processing (EMNLP)}, pages 2255--2266, Online. Association
  for Computational Linguistics.

\bibitem[{Khayrallah and Koehn(2018)}]{khayrallah-koehn-2018-impact}
Huda Khayrallah and Philipp Koehn. 2018.
\newblock \href {https://doi.org/10.18653/v1/W18-2709} {On the impact of
  various types of noise on neural machine translation}.
\newblock In \emph{Proceedings of the 2nd Workshop on Neural Machine
  Translation and Generation}, pages 74--83, Melbourne, Australia. Association
  for Computational Linguistics.

\bibitem[{Kim and Rush(2016)}]{kim-rush-2016-sequence}
Yoon Kim and Alexander~M. Rush. 2016.
\newblock \href {https://doi.org/10.18653/v1/D16-1139} {Sequence-level
  knowledge distillation}.
\newblock In \emph{Proceedings of the 2016 Conference on Empirical Methods in
  Natural Language Processing}, pages 1317--1327, Austin, Texas. Association
  for Computational Linguistics.

\bibitem[{Koehn et~al.(2020)Koehn, Chaudhary, El-Kishky, Goyal, Chen, and
  Guzm{\'a}n}]{koehn-etal-2020-findings}
Philipp Koehn, Vishrav Chaudhary, Ahmed El-Kishky, Naman Goyal, Peng-Jen Chen,
  and Francisco Guzm{\'a}n. 2020.
\newblock \href {https://aclanthology.org/2020.wmt-1.78} {Findings of the {WMT}
  2020 shared task on parallel corpus filtering and alignment}.
\newblock In \emph{Proceedings of the Fifth Conference on Machine Translation},
  pages 726--742, Online. Association for Computational Linguistics.

\bibitem[{Koehn et~al.(2019)Koehn, Guzm{\'a}n, Chaudhary, and
  Pino}]{koehn-etal-2019-findings}
Philipp Koehn, Francisco Guzm{\'a}n, Vishrav Chaudhary, and Juan Pino. 2019.
\newblock \href {https://doi.org/10.18653/v1/W19-5404} {Findings of the {WMT}
  2019 shared task on parallel corpus filtering for low-resource conditions}.
\newblock In \emph{Proceedings of the Fourth Conference on Machine Translation
  (Volume 3: Shared Task Papers, Day 2)}, pages 54--72, Florence, Italy.
  Association for Computational Linguistics.

\bibitem[{Koehn et~al.(2007)Koehn, Hoang, Birch, Callison-Burch, Federico,
  Bertoldi, Cowan, Shen, Moran, Zens, Dyer, Bojar, Constantin, and
  Herbst}]{koehn-etal-2007-moses}
Philipp Koehn, Hieu Hoang, Alexandra Birch, Chris Callison-Burch, Marcello
  Federico, Nicola Bertoldi, Brooke Cowan, Wade Shen, Christine Moran, Richard
  Zens, Chris Dyer, Ond{\v{r}}ej Bojar, Alexandra Constantin, and Evan Herbst.
  2007.
\newblock \href {https://aclanthology.org/P07-2045} {{M}oses: Open source
  toolkit for statistical machine translation}.
\newblock In \emph{Proceedings of the 45th Annual Meeting of the Association
  for Computational Linguistics Companion Volume Proceedings of the Demo and
  Poster Sessions}, pages 177--180, Prague, Czech Republic. Association for
  Computational Linguistics.

\bibitem[{Koehn et~al.(2018)Koehn, Khayrallah, Heafield, and
  Forcada}]{koehn-etal-2018-findings}
Philipp Koehn, Huda Khayrallah, Kenneth Heafield, and Mikel~L. Forcada. 2018.
\newblock \href {https://doi.org/10.18653/v1/W18-6453} {Findings of the {WMT}
  2018 shared task on parallel corpus filtering}.
\newblock In \emph{Proceedings of the Third Conference on Machine Translation:
  Shared Task Papers}, pages 726--739, Belgium, Brussels. Association for
  Computational Linguistics.

\bibitem[{Koehn and Knowles(2017)}]{koehn-knowles-2017-six}
Philipp Koehn and Rebecca Knowles. 2017.
\newblock \href {https://doi.org/10.18653/v1/W17-3204} {Six challenges for
  neural machine translation}.
\newblock In \emph{Proceedings of the First Workshop on Neural Machine
  Translation}, pages 28--39, Vancouver. Association for Computational
  Linguistics.

\bibitem[{Kreutzer et~al.(2022)Kreutzer, Caswell, Wang, Wahab, van Esch,
  Ulzii-Orshikh, Tapo, Subramani, Sokolov, Sikasote, Setyawan, Sarin, Samb,
  Sagot, Rivera, Rios, Papadimitriou, Osei, Suarez, Orife, Ogueji, Rubungo,
  Nguyen, Müller, Müller, Muhammad, Muhammad, Mnyakeni, Mirzakhalov,
  Matangira, Leong, Lawson, Kudugunta, Jernite, Jenny, Firat, Dossou, Dlamini,
  de~Silva, Çabuk Ballı, Biderman, Battisti, Baruwa, Bapna, Baljekar, Azime,
  Awokoya, Ataman, Ahia, Ahia, Agrawal, and Adeyemi}]{mashakane}
Julia Kreutzer, Isaac Caswell, Lisa Wang, Ahsan Wahab, Daan van Esch,
  Nasanbayar Ulzii-Orshikh, Allahsera Tapo, Nishant Subramani, Artem Sokolov,
  Claytone Sikasote, Monang Setyawan, Supheakmungkol Sarin, Sokhar Samb,
  Benoît Sagot, Clara Rivera, Annette Rios, Isabel Papadimitriou, Salomey
  Osei, Pedro~Ortiz Suarez, Iroro Orife, Kelechi Ogueji, Andre~Niyongabo
  Rubungo, Toan~Q. Nguyen, Mathias Müller, André Müller, Shamsuddeen~Hassan
  Muhammad, Nanda Muhammad, Ayanda Mnyakeni, Jamshidbek Mirzakhalov,
  Tapiwanashe Matangira, Colin Leong, Nze Lawson, Sneha Kudugunta, Yacine
  Jernite, Mathias Jenny, Orhan Firat, Bonaventure F.~P. Dossou, Sakhile
  Dlamini, Nisansa de~Silva, Sakine Çabuk Ballı, Stella Biderman, Alessia
  Battisti, Ahmed Baruwa, Ankur Bapna, Pallavi Baljekar, Israel~Abebe Azime,
  Ayodele Awokoya, Duygu Ataman, Orevaoghene Ahia, Oghenefego Ahia, Sweta
  Agrawal, and Mofetoluwa Adeyemi. 2022.
\newblock \href {https://doi.org/10.1162/tacl_a_00447} {{Quality at a Glance:
  An Audit of Web-Crawled Multilingual Datasets}}.
\newblock \emph{Transactions of the Association for Computational Linguistics},
  10:50--72.

\bibitem[{Lison and Tiedemann(2016)}]{lison-tiedemann-2016-opensubtitles2016}
Pierre Lison and J{\"o}rg Tiedemann. 2016.
\newblock \href {https://aclanthology.org/L16-1147} {{O}pen{S}ubtitles2016:
  Extracting large parallel corpora from movie and {TV} subtitles}.
\newblock In \emph{Proceedings of the Tenth International Conference on
  Language Resources and Evaluation ({LREC}'16)}, pages 923--929,
  Portoro{\v{z}}, Slovenia. European Language Resources Association (ELRA).

\bibitem[{Liu et~al.(2020)Liu, Gu, Goyal, Li, Edunov, Ghazvininejad, Lewis, and
  Zettlemoyer}]{liu-etal-2020-multilingual-denoising}
Yinhan Liu, Jiatao Gu, Naman Goyal, Xian Li, Sergey Edunov, Marjan
  Ghazvininejad, Mike Lewis, and Luke Zettlemoyer. 2020.
\newblock \href {https://doi.org/10.1162/tacl_a_00343} {Multilingual denoising
  pre-training for neural machine translation}.
\newblock \emph{Transactions of the Association for Computational Linguistics},
  8:726--742.

\bibitem[{Lopes et~al.(2019)Lopes, Farajian, Correia, Tr{\'e}nous, and
  Martins}]{lopes-etal-2019-unbabels}
Ant{\'o}nio~V. Lopes, M.~Amin Farajian, Gon{\c{c}}alo~M. Correia, Jonay
  Tr{\'e}nous, and Andr{\'e} F.~T. Martins. 2019.
\newblock \href {https://doi.org/10.18653/v1/W19-5413} {Unbabel{'}s submission
  to the {WMT}2019 {APE} shared task: {BERT}-based encoder-decoder for
  automatic post-editing}.
\newblock In \emph{Proceedings of the Fourth Conference on Machine Translation
  (Volume 3: Shared Task Papers, Day 2)}, pages 118--123, Florence, Italy.
  Association for Computational Linguistics.

\bibitem[{Lu et~al.(2018)Lu, Lv, Shi, and Chen}]{lu-etal-2018-alibaba}
Jun Lu, Xiaoyu Lv, Yangbin Shi, and Boxing Chen. 2018.
\newblock \href {https://doi.org/10.18653/v1/W18-6481} {{A}libaba submission to
  the {WMT}18 parallel corpus filtering task}.
\newblock In \emph{Proceedings of the Third Conference on Machine Translation:
  Shared Task Papers}, pages 917--922, Belgium, Brussels. Association for
  Computational Linguistics.

\bibitem[{Marie et~al.(2020)Marie, Rubino, and Fujita}]{marie-etal-2020-tagged}
Benjamin Marie, Raphael Rubino, and Atsushi Fujita. 2020.
\newblock \href {https://doi.org/10.18653/v1/2020.acl-main.532} {Tagged
  back-translation revisited: Why does it really work?}
\newblock In \emph{Proceedings of the 58th Annual Meeting of the Association
  for Computational Linguistics}, pages 5990--5997, Online. Association for
  Computational Linguistics.

\bibitem[{Nguyen et~al.(2020)Nguyen, Joty, Kui, and
  Aw}]{10.5555/3495724.3496564}
Xuan-Phi Nguyen, Shafiq Joty, Wu~Kui, and Ai~Ti Aw. 2020.
\newblock Data diversification: A simple strategy for neural machine
  translation.
\newblock In \emph{Proceedings of the 34th International Conference on Neural
  Information Processing Systems}, NIPS'20, Red Hook, NY, USA. Curran
  Associates Inc.

\bibitem[{Ott et~al.(2018)Ott, Auli, Grangier, and Ranzato}]{pmlr-v80-ott18a}
Myle Ott, Michael Auli, David Grangier, and Marc'Aurelio Ranzato. 2018.
\newblock \href {https://proceedings.mlr.press/v80/ott18a.html} {Analyzing
  uncertainty in neural machine translation}.
\newblock In \emph{Proceedings of the 35th International Conference on Machine
  Learning}, volume~80 of \emph{Proceedings of Machine Learning Research},
  pages 3956--3965. PMLR.

\bibitem[{Ott et~al.(2019)Ott, Edunov, Baevski, Fan, Gross, Ng, Grangier, and
  Auli}]{ott2019fairseq}
Myle Ott, Sergey Edunov, Alexei Baevski, Angela Fan, Sam Gross, Nathan Ng,
  David Grangier, and Michael Auli. 2019.
\newblock \href {https://doi.org/10.18653/v1/N19-4009} {fairseq: A fast,
  extensible toolkit for sequence modeling}.
\newblock In \emph{Proceedings of the 2019 Conference of the North {A}merican
  Chapter of the Association for Computational Linguistics (Demonstrations)},
  pages 48--53, Minneapolis, Minnesota. Association for Computational
  Linguistics.

\bibitem[{Papavassiliou et~al.(2018)Papavassiliou, Sofianopoulos, Prokopidis,
  and Piperidis}]{papavassiliou-etal-2018-ilsp}
Vassilis Papavassiliou, Sokratis Sofianopoulos, Prokopis Prokopidis, and
  Stelios Piperidis. 2018.
\newblock \href {https://doi.org/10.18653/v1/W18-6484} {The {ILSP}/{ARC}
  submission to the {WMT} 2018 parallel corpus filtering shared task}.
\newblock In \emph{Proceedings of the Third Conference on Machine Translation:
  Shared Task Papers}, pages 928--933, Belgium, Brussels. Association for
  Computational Linguistics.

\bibitem[{Popovi{\'c}(2015)}]{popovic-2015-chrf}
Maja Popovi{\'c}. 2015.
\newblock \href {https://doi.org/10.18653/v1/W15-3049} {chr{F}: character
  n-gram {F}-score for automatic {MT} evaluation}.
\newblock In \emph{Proceedings of the Tenth Workshop on Statistical Machine
  Translation}, pages 392--395, Lisbon, Portugal. Association for Computational
  Linguistics.

\bibitem[{Resnik(1999)}]{resnik-1999-mining}
Philip Resnik. 1999.
\newblock \href {https://doi.org/10.3115/1034678.1034757} {Mining the web for
  bilingual text}.
\newblock In \emph{Proceedings of the 37th Annual Meeting of the Association
  for Computational Linguistics}, pages 527--534, College Park, Maryland, USA.
  Association for Computational Linguistics.

\bibitem[{Rossenbach et~al.(2018)Rossenbach, Rosendahl, Kim, Gra{\c{c}}a,
  Gokrani, and Ney}]{rossenbach-etal-2018-rwth}
Nick Rossenbach, Jan Rosendahl, Yunsu Kim, Miguel Gra{\c{c}}a, Aman Gokrani,
  and Hermann Ney. 2018.
\newblock \href {https://doi.org/10.18653/v1/W18-6487} {The {RWTH} {A}achen
  {U}niversity filtering system for the {WMT} 2018 parallel corpus filtering
  task}.
\newblock In \emph{Proceedings of the Third Conference on Machine Translation:
  Shared Task Papers}, pages 946--954, Belgium, Brussels. Association for
  Computational Linguistics.

\bibitem[{S{\'a}nchez-Cartagena et~al.(2018)S{\'a}nchez-Cartagena,
  Ba{\~n}{\'o}n, Ortiz-Rojas, and
  Ram{\'\i}rez}]{sanchez-cartagena-etal-2018-prompsits}
V{\'\i}ctor~M. S{\'a}nchez-Cartagena, Marta Ba{\~n}{\'o}n, Sergio Ortiz-Rojas,
  and Gema Ram{\'\i}rez. 2018.
\newblock \href {https://doi.org/10.18653/v1/W18-6488} {Prompsit{'}s submission
  to {WMT} 2018 parallel corpus filtering shared task}.
\newblock In \emph{Proceedings of the Third Conference on Machine Translation:
  Shared Task Papers}, pages 955--962, Belgium, Brussels. Association for
  Computational Linguistics.

\bibitem[{Schwenk(2018)}]{schwenk-2018-filtering}
Holger Schwenk. 2018.
\newblock \href {https://doi.org/10.18653/v1/P18-2037} {Filtering and mining
  parallel data in a joint multilingual space}.
\newblock In \emph{Proceedings of the 56th Annual Meeting of the Association
  for Computational Linguistics (Volume 2: Short Papers)}, pages 228--234,
  Melbourne, Australia. Association for Computational Linguistics.

\bibitem[{Schwenk et~al.(2021{\natexlab{a}})Schwenk, Chaudhary, Sun, Gong, and
  Guzm{\'a}n}]{schwenk-etal-2021-wikimatrix}
Holger Schwenk, Vishrav Chaudhary, Shuo Sun, Hongyu Gong, and Francisco
  Guzm{\'a}n. 2021{\natexlab{a}}.
\newblock \href {https://aclanthology.org/2021.eacl-main.115} {{W}iki{M}atrix:
  Mining 135{M} parallel sentences in 1620 language pairs from {W}ikipedia}.
\newblock In \emph{Proceedings of the 16th Conference of the European Chapter
  of the Association for Computational Linguistics: Main Volume}, pages
  1351--1361, Online. Association for Computational Linguistics.

\bibitem[{Schwenk et~al.(2021{\natexlab{b}})Schwenk, Wenzek, Edunov, Grave,
  Joulin, and Fan}]{schwenk-etal-2021-ccmatrix}
Holger Schwenk, Guillaume Wenzek, Sergey Edunov, Edouard Grave, Armand Joulin,
  and Angela Fan. 2021{\natexlab{b}}.
\newblock \href {https://doi.org/10.18653/v1/2021.acl-long.507} {{CCM}atrix:
  Mining billions of high-quality parallel sentences on the web}.
\newblock In \emph{Proceedings of the 59th Annual Meeting of the Association
  for Computational Linguistics and the 11th International Joint Conference on
  Natural Language Processing (Volume 1: Long Papers)}, pages 6490--6500,
  Online. Association for Computational Linguistics.

\bibitem[{Sennrich et~al.(2016{\natexlab{a}})Sennrich, Haddow, and
  Birch}]{sennrich-etal-2016-improving}
Rico Sennrich, Barry Haddow, and Alexandra Birch. 2016{\natexlab{a}}.
\newblock \href {https://doi.org/10.18653/v1/P16-1009} {Improving neural
  machine translation models with monolingual data}.
\newblock In \emph{Proceedings of the 54th Annual Meeting of the Association
  for Computational Linguistics (Volume 1: Long Papers)}, pages 86--96, Berlin,
  Germany. Association for Computational Linguistics.

\bibitem[{Sennrich et~al.(2016{\natexlab{b}})Sennrich, Haddow, and
  Birch}]{sennrich-etal-2016-neural}
Rico Sennrich, Barry Haddow, and Alexandra Birch. 2016{\natexlab{b}}.
\newblock \href {https://doi.org/10.18653/v1/P16-1162} {Neural machine
  translation of rare words with subword units}.
\newblock In \emph{Proceedings of the 54th Annual Meeting of the Association
  for Computational Linguistics (Volume 1: Long Papers)}, pages 1715--1725,
  Berlin, Germany. Association for Computational Linguistics.

\bibitem[{Snover et~al.(2006)Snover, Dorr, Schwartz, Micciulla, and
  Makhoul}]{snover-etal-2006-study}
Matthew Snover, Bonnie Dorr, Rich Schwartz, Linnea Micciulla, and John Makhoul.
  2006.
\newblock \href {https://aclanthology.org/2006.amta-papers.25} {A study of
  translation edit rate with targeted human annotation}.
\newblock In \emph{Proceedings of the 7th Conference of the Association for
  Machine Translation in the Americas: Technical Papers}, pages 223--231,
  Cambridge, Massachusetts, USA. Association for Machine Translation in the
  Americas.

\bibitem[{Tiedemann(2020)}]{tiedemann-2020-tatoeba}
J{\"o}rg Tiedemann. 2020.
\newblock \href {https://aclanthology.org/2020.wmt-1.139} {The tatoeba
  translation challenge {--} realistic data sets for low resource and
  multilingual {MT}}.
\newblock In \emph{Proceedings of the Fifth Conference on Machine Translation},
  pages 1174--1182, Online. Association for Computational Linguistics.

\bibitem[{Tiedemann(2012)}]{TIEDEMANN12.463}
Jörg Tiedemann. 2012.
\newblock Parallel data, tools and interfaces in opus.
\newblock In \emph{Proceedings of the Eight International Conference on
  Language Resources and Evaluation (LREC'12)}, Istanbul, Turkey. European
  Language Resources Association (ELRA).

\bibitem[{Tran et~al.(2020)Tran, Tang, Li, and Gu}]{NEURIPS2020_1763ea5a}
Chau Tran, Yuqing Tang, Xian Li, and Jiatao Gu. 2020.
\newblock \href
  {https://proceedings.neurips.cc/paper/2020/file/1763ea5a7e72dd7ee64073c2dda7a7a8-Paper.pdf}
  {Cross-lingual retrieval for iterative self-supervised training}.
\newblock In \emph{Advances in Neural Information Processing Systems},
  volume~33, pages 2207--2219. Curran Associates, Inc.

\bibitem[{Vanmassenhove et~al.(2019)Vanmassenhove, Shterionov, and
  Way}]{vanmassenhove-etal-2019-lost}
Eva Vanmassenhove, Dimitar Shterionov, and Andy Way. 2019.
\newblock \href {https://aclanthology.org/W19-6622} {Lost in translation: Loss
  and decay of linguistic richness in machine translation}.
\newblock In \emph{Proceedings of Machine Translation Summit XVII: Research
  Track}, pages 222--232, Dublin, Ireland. European Association for Machine
  Translation.

\bibitem[{Vaswani et~al.(2017)Vaswani, Shazeer, Parmar, Uszkoreit, Jones,
  Gomez, Kaiser, and Polosukhin}]{NIPS2017_3f5ee243}
Ashish Vaswani, Noam Shazeer, Niki Parmar, Jakob Uszkoreit, Llion Jones,
  Aidan~N Gomez, \L~ukasz Kaiser, and Illia Polosukhin. 2017.
\newblock \href
  {https://proceedings.neurips.cc/paper/2017/file/3f5ee243547dee91fbd053c1c4a845aa-Paper.pdf}
  {Attention is all you need}.
\newblock In \emph{Advances in Neural Information Processing Systems},
  volume~30. Curran Associates, Inc.

\bibitem[{Vyas et~al.(2018)Vyas, Niu, and Carpuat}]{vyas-etal-2018-identifying}
Yogarshi Vyas, Xing Niu, and Marine Carpuat. 2018.
\newblock \href {https://doi.org/10.18653/v1/N18-1136} {Identifying semantic
  divergences in parallel text without annotations}.
\newblock In \emph{Proceedings of the 2018 Conference of the North {A}merican
  Chapter of the Association for Computational Linguistics: Human Language
  Technologies, Volume 1 (Long Papers)}, pages 1503--1515, New Orleans,
  Louisiana. Association for Computational Linguistics.

\bibitem[{Wei et~al.(2020)Wei, Zhang, Chen, and Luo}]{wei-etal-2020-iterative}
Hao-Ran Wei, Zhirui Zhang, Boxing Chen, and Weihua Luo. 2020.
\newblock \href {https://doi.org/10.18653/v1/2020.emnlp-main.474} {Iterative
  domain-repaired back-translation}.
\newblock In \emph{Proceedings of the 2020 Conference on Empirical Methods in
  Natural Language Processing (EMNLP)}, pages 5884--5893, Online. Association
  for Computational Linguistics.

\bibitem[{Weston et~al.(2018)Weston, Dinan, and
  Miller}]{weston-etal-2018-retrieve}
Jason Weston, Emily Dinan, and Alexander Miller. 2018.
\newblock \href {https://doi.org/10.18653/v1/W18-5713} {Retrieve and refine:
  Improved sequence generation models for dialogue}.
\newblock In \emph{Proceedings of the 2018 {EMNLP} Workshop {SCAI}: The 2nd
  International Workshop on Search-Oriented Conversational {AI}}, pages 87--92,
  Brussels, Belgium. Association for Computational Linguistics.

\bibitem[{Yang et~al.(2020)Yang, Wang, Wei, Shang, Guo, Li, Lei, Qin, Tao, Sun,
  and Chen}]{yang-etal-2020-hw}
Hao Yang, Minghan Wang, Daimeng Wei, Hengchao Shang, Jiaxin Guo, Zongyao Li,
  Lizhi Lei, Ying Qin, Shimin Tao, Shiliang Sun, and Yimeng Chen. 2020.
\newblock \href {https://aclanthology.org/2020.wmt-1.85} {{HW}-{TSC}{'}s
  participation at {WMT} 2020 automatic post editing shared task}.
\newblock In \emph{Proceedings of the Fifth Conference on Machine Translation},
  pages 797--802, Online. Association for Computational Linguistics.

\bibitem[{Zhang and Zong(2016)}]{zhang-zong-2016-exploiting}
Jiajun Zhang and Chengqing Zong. 2016.
\newblock \href {https://doi.org/10.18653/v1/D16-1160} {Exploiting source-side
  monolingual data in neural machine translation}.
\newblock In \emph{Proceedings of the 2016 Conference on Empirical Methods in
  Natural Language Processing}, pages 1535--1545, Austin, Texas. Association
  for Computational Linguistics.

\bibitem[{Zhang et~al.(2018)Zhang, Utiyama, Sumita, Neubig, and
  Nakamura}]{zhang-etal-2018-guiding}
Jingyi Zhang, Masao Utiyama, Eiichro Sumita, Graham Neubig, and Satoshi
  Nakamura. 2018.
\newblock \href {https://doi.org/10.18653/v1/N18-1120} {Guiding neural machine
  translation with retrieved translation pieces}.
\newblock In \emph{Proceedings of the 2018 Conference of the North {A}merican
  Chapter of the Association for Computational Linguistics: Human Language
  Technologies, Volume 1 (Long Papers)}, pages 1325--1335, New Orleans,
  Louisiana. Association for Computational Linguistics.

\bibitem[{Zhou et~al.(2021)Zhou, Neubig, Gu, Diab, Guzm{\'a}n, Zettlemoyer, and
  Ghazvininejad}]{zhou-etal-2021-detecting}
Chunting Zhou, Graham Neubig, Jiatao Gu, Mona Diab, Francisco Guzm{\'a}n, Luke
  Zettlemoyer, and Marjan Ghazvininejad. 2021.
\newblock \href {https://doi.org/10.18653/v1/2021.findings-acl.120} {Detecting
  hallucinated content in conditional neural sequence generation}.
\newblock In \emph{Findings of the Association for Computational Linguistics:
  ACL-IJCNLP 2021}, pages 1393--1404, Online. Association for Computational
  Linguistics.

\bibitem[{Zoph et~al.(2016)Zoph, Yuret, May, and
  Knight}]{zoph-etal-2016-transfer}
Barret Zoph, Deniz Yuret, Jonathan May, and Kevin Knight. 2016.
\newblock \href {https://doi.org/10.18653/v1/D16-1163} {Transfer learning for
  low-resource neural machine translation}.
\newblock In \emph{Proceedings of the 2016 Conference on Empirical Methods in
  Natural Language Processing}, pages 1568--1575, Austin, Texas. Association
  for Computational Linguistics.

\end{thebibliography}
\bibliographystyle{acl_natbib}

\clearpage
\appendix

\section{Results on Second Evaluation Metric}\label{sec:chrf}
Table~\ref{sec:chrf} presents results on \textsc{nmt} tasks for a second evaluation metric. 
\begin{table}[!ht]
    \centering
    \scalebox{0.75}{
    \begin{tabular}{cccccc}
     \toprule
            & \enoc & \itoc & \enbe & \enmr & \ensw \\
        \cmidrule(lr){2-6}

        1 : &  $41.59$ & $30.92$ & $29.28$ & $31.19$ & $59.17$ \\
        2 : & $39.73$  & $32.26$ & $28.24$ & $31.90$ & $58.76$ \\
        3 : & $42.34$  & $40.62$ & $30.96$ & $35.41$ & $58.60$ \\
        4 : & $47.40$  & $42.83$ & $31.21$ & $35.01$ & $59.02$ \\
        5 : & $44.66$  & $39.01$ & $30.66$ & $35.20$ & $59.50$ \\
        6 : & $47.74$  & $43.03$ & $31.08$ & $34.65$ & $59.49$ \\
        
        \addlinespace[0.3cm]
        \hline
        \addlinespace[0.3cm]
 
        & \ocen & \ocit & \been & \mren & \swen \\
        \cmidrule(lr){2-6}
        7 : & $48.04$  & $32.90$ & $37.13$ & $37.84$ & $57.10$ \\
        8 : & $42.10$  & $33.73$ & $33.51$ & $36.55$ & $57.07$ \\
        9 : & $50.99$  & $42.42$ & $37.13$ & $39.99$ & $56.74$ \\
        10 : & $52.13$ & $42.42$ & $39.20$ & $42.45$ & $57.96$ \\
        11 : & $51.63$ & $38.99$ & $36.99$ & $40.29$ & $58.74$ \\
        12 : & $53.86$ & $44.05$ & $39.18$ & $42.71$ & $58.29$ \\
        \bottomrule
    \end{tabular}}
    \caption{Results on \textsc{nmt} tasks for the chrF metric (rows follow the enumeration of Table~\ref{tab:main_results}).}
    \label{tab:chrf}
\end{table}

\section{Scaling-Up Settings}\label{sec:analysis_details}
Tables~\ref{tab:analysis_setting_a}, \ref{tab:analysis_setting_b}, and \ref{tab:analysis_setting_c} present training data sizes for experiments in Figure~\ref{fig:scaling_up}.
\begin{table}[!ht]
    \centering
    \scalebox{0.9}{
    \begin{tabular}{lrrrrr}
    \toprule
    A   & $0.5$M & $1.0$M & $5.0$M & $10.0$M & $15.0$M\\
    A$\cup$B  & $0.75$M & $1.5$M & $7.5$M & $15.0$M & $22.5$M \\
    A$\cup$r(B) & $0.75$M & $1.5$M & $7.5$M & $15.0$M & $22.5$M \\
    \bottomrule
    \end{tabular}}
    \caption{Training data size for experiments in Figure~\ref{fig:scaling_up}(a), where $|B| = |A|/2 $.}
    \label{tab:analysis_setting_a}
\end{table}
\begin{table}[!ht]
    \centering
    \scalebox{0.9}{
    \begin{tabular}{lrrrrr}
    \toprule
    A   & $0.5$M & $1.0$M & $5.0$M & $10.0$M & $15.0$M\\
    A$\cup$B  & $1.0$M & $2.0$M & $10.0$M & $20.0$M & $30.0$M\\
    A$\cup$r(B) & $1.0$M & $2.0$M & $10.0$M & $20.0$M & $30.0$M\\
    \bottomrule
    \end{tabular}}
    \caption{Training data size for experiments in Figure~\ref{fig:scaling_up}(b), where $|B| = |A| $.}
    \label{tab:analysis_setting_b}
\end{table}
\begin{table}[!ht]
    \centering
    \scalebox{0.9}{
    \begin{tabular}{lrrrrr}
    \toprule
    A   & $0.5$M & $1.0$M & $5.0$M & $10.0$M & $15.0$M\\
    A$\cup$B  & $1.5$M & $3.0$M & $15.0$M & $30.0M$ & $70$M\\
    A$\cup$r(B) & $1.0$M & $2.0$M & $10.0$M & $20.0$M & $70.0$M\\
    \bottomrule
    \end{tabular}}
    \caption{Training data size for experiments in Figure~\ref{fig:scaling_up}(c), where $|B| = 2|A| $.}
    \label{tab:analysis_setting_c}
\end{table}

\section{Manual Annotation Details}\label{sec:manual_annotation_details}

For each bitext (i.e., original CCMatrix sample or refined sample edited by a bitext refinement model) we rate the \textbf{degree of equivalence} between the two sentences following the protocol of semantic divergences~\cite{briakou-carpuat-2020-detecting}. Concretely, a bitext is annotated as having \textit{no meaning difference} if it corresponds to perfect translations, \textit{some meaning differences} if the sentences share important content in common but differ by few tokens (e.g., small added content, or phrasal mistranslation), and \textit{unrelated} if the sentences are only topically or structurally related. For rating \textbf{fluency} we evaluate the output sentence of the bitext refinement models in isolation on a discrete scale of 1 to 5, following \citet{heilman-etal-2014-predicting} (Other → Incomprehensible → Somewhat
Comprehensible → Comprehensible → Perfect).

\section{Fairseq configuration details}\label{sec:fairseq_settings}
Table~\ref{tab:fairseq} presents details of \textsc{nmt} training with fairseq. The same parameters are used to train \textsc{BitextEdit} models. 
\begin{table}[!ht]
    \centering
    \scalebox{0.8}{
    \begin{tabular}{|l|}
    \hline
       \texttt{--arch transformer} \\
       \texttt{--share-all-embeddings} \\
       \texttt{--encoder-layers 6} \\
       \texttt{--decoder-layers 6} \\
       \texttt{--encoder-embed-dim 512} \\
       \texttt{--decoder-embed-dim 512} \\
       \texttt{--encoder-ffn-embed-dim 4096} \\
       \texttt{--decoder-ffn-embed-dim 4096} \\
       \texttt{--encoder-attention-heads 8}  \\
       \texttt{--decoder-attention-heads 8} \\
       \texttt{--encoder-normalize-before} \\
       \texttt{--decoder-normalize-before} \\
       \texttt{--dropout 0.4} \\
       \texttt{--attention-dropout 0.2} \\
       \texttt{--relu-dropout 0.2} \\
       \texttt{--weight-decay 0.0001} \\
       \texttt{--label-smoothing 0.2} \\
       \texttt{--criterion label smoothed cross entropy} \\
       \texttt{--optimizer adam} \\
       \texttt{--adam-betas '(0.9, 0.98)'} \\
       \texttt{--clip-norm 0} \\
       \texttt{--lr-scheduler inverse sqrt} \\
       \texttt{--warmup-updates 4000} \\
       \texttt{--warmup-init-lr 1e-7}\\
       \texttt{--lr 1e-3} \\
       \texttt{--max-tokens 4000} \\
       \texttt{--update-freq 4} \\
       \texttt{--max-epoch 100} \\
       \texttt{--save-interval 10} \\
    \hline
    \end{tabular}}\vspace{-0.5em}
    \caption{Fairseq configuration used for \textsc{nmt} training.}
    \label{tab:fairseq}
\end{table}

\section{\textsc{BitextEdit:} Model outputs}\label{sec:model_outputs}
Table~\ref{tab:model_outputs} presents model outputs samples edited by our model for \textsc{en-el} CCMatrix instances.

\begin{table*}[!ht]
    \centering
    \scalebox{0.7}{
    \begin{tabular}{lll}
    %
    \toprule
    \addlinespace[1.0em]

    \editedside & \textsc{[en]} \small{\textsc{ccmatrix}} & Respect the dignity of all people, regardless of their age. \\
    & \textsc{[el]} \small{\textsc{ccmatrix}}  &
    \selectlanguage{greek}
    \colorbox{red!10}{\hz Πιστεύω στην αναγκαιότητα αξιοποίησης όλων των άξιων ανθρώπων} ανεξάρτητα από την ηλικία τους.  \\
    & \hspace{0.5em} $\lfloor$ \hspace{0.3em} \small{\textsc{gloss}}  &
    {\color{gray!80}
    \hspace{0.1em}
    \textit{
    I believe in the importance of using all skilled people, regardless of their age.}}
    \\
    & \textsc{[el]} \small{\textsc{BitextEdit}} & 
    \selectlanguage{greek}
    Σεβασμός στην αξιοπρέπεια όλων των ανθρώπων , ανεξάρτητα από την ηλικία τους. \\
    \addlinespace[1.0em]
    \cmidrule{2-3}
    \addlinespace[1.0em]
 
    %
    \editedside & \textsc{[en]} \small{\textsc{ccmatrix}} & \colorbox{red!10}{\hz After that time} the \colorbox{red!10}{\hz whole} group would \colorbox{red!10}{\hz talk} for 5 minutes. \\
    & \textsc{[el]} \small{\textsc{ccmatrix}} & 
    \selectlanguage{greek}
    Αργότερα, η ομάδα μελέτης ζήτησε από όλους να διαλογιστούν για πέντε λεπτά. \\
    & \hspace{0.5em} $\lfloor$ \hspace{0.3em} \small{\textsc{gloss}}  &
    {\color{gray!80}
    \hspace{0.1em}
    \textit{
    Later, the study group asked everyone to meditate for 5 minutes.}}\\
    & \textsc{[en]} \small{\textsc{BitextEdit}} & 
    Later, the study group asked everyone to meditate for five minutes. \\

    \addlinespace[1.0em]
    \cmidrule{2-3}
    \addlinespace[1.0em]

    & \textsc{[en]} \small{\textsc{ccmatrix}} & Say no to fake products and scams. \\
    \editedside & \textsc{[el]} \small{\textsc{ccmatrix}} & 
    \selectlanguage{greek} 
    \colorbox{red!10}{\hz Είπατε} όχι στις ψεύτικες υποσχέσεις και στη συναλλαγή. \\
    & \hspace{0.5em} $\lfloor$ \hspace{0.3em} \small{\textsc{gloss}}  &
    {\color{gray!80}
    \hspace{0.1em}
    \textit{
    You said no to fake products and transactions.}}\\
    & \textsc{[el]} \small{\textsc{BitextEdit}} & 
    \selectlanguage{greek}
    Πείτε όχι στα ψεύτικα προϊόντα και απάτες. \\
    \addlinespace[1.0em]
    \cmidrule{2-3}
    \addlinespace[1.0em]

    \editedside & \textsc{[en]} \small{\textsc{ccmatrix}} & 
    \colorbox{red!10}{\hz We’re all} part of a larger system. \\
    & \textsc{[el]} \small{\textsc{ccmatrix}} & 
    \selectlanguage{greek}
    Τα πάντα είναι μέρος ενός μεγαλύτερου Συστήματος. \\
    & \hspace{0.5em} $\lfloor$ \hspace{0.3em} \small{\textsc{gloss}}  &
    {\color{gray!80}
    \hspace{0.1em}
    \textit{
    Everying is part of a larger System.}}\\    %
    & \textsc{[en]} \small{\textsc{BitextEdit}} & 
    Everything is part of a larger system. \\
    \addlinespace[1.0em]
    \cmidrule{2-3}
    \addlinespace[1.0em]

    \editedside & \textsc{[en]} \small{\textsc{ccmatrix}} & 
    Currently, no \colorbox{red!10}{\hz equivalent} technology exists on the market . \\
    & \textsc{[en]} \small{\textsc{ccmatrix}} &
    \selectlanguage{greek}
    Δεν υπάρχει αντίστοιχη ανταγωνιστική τεχνολογία στον κόσμο αυτή τη στιγμή. \\
     & \hspace{0.5em} $\lfloor$ \hspace{0.3em} \small{\textsc{gloss}}  &
    {\color{gray!80}
    \hspace{0.1em}
    \textit{
    There is no corresponding competing technology in the word right now.}}\\
    & \textsc{[en]} \small{\textsc{BitextEdit}} & 
    There is no \colorbox{red!10}{\hz competitive} technology in the world right now. \\
    \addlinespace[1.0em]
    \cmidrule{2-3}
    \addlinespace[1.0em]

    & \textsc{[en]} \small{\textsc{ccmatrix}} & 
    ``The western influence came from film and television'', he later explained. \\
    \editedside & \textsc{[en]} \small{\textsc{ccmatrix}} &
    \selectlanguage{greek} 
    «Η \colorbox{red!10}{\hz λογοκρισία εντείνεται όλο και περισσότερο στον κινηματογράφο} και την τηλεόραση», εξήγησε ο ίδιος. \\
     & \hspace{0.5em} $\lfloor$ \hspace{0.3em} \small{\textsc{gloss}}  &
    {\color{gray!80}
    \hspace{0.1em}
    \textit{
    ``Censorship is intensifying in cinema and television'', he explained.
    }}\\
    & \textsc{[el]} \small{\textsc{BitextEdit}} & 
    \selectlanguage{greek}
     «Η δυτική επιρροή ήρθε από την ταινία και την τηλεόραση» , εξήγησε αργότερα. \\
    \addlinespace[1.0em]
    \cmidrule{2-3}
    \addlinespace[1.0em]

     & \textsc{[en]} \small{\textsc{ccmatrix}} & 
     Then he paused, surveying the surreal scene. \\
     \editedside & \textsc{[en]} \small{\textsc{ccmatrix}} & 
     \selectlanguage{greek}
     Και πράγματι \colorbox{red!10}{\hz έφυγε}, \colorbox{red!10}{\hz προσπερνώντας τον έκπληκτο Κέλι}. \\
     & \hspace{0.5em} $\lfloor$ \hspace{0.3em} \small{\textsc{gloss}}  &
    {\color{gray!80}
    \hspace{0.1em}
    \textit{
    And indeed he left, passing Keli, who was surprised.
    }}\\    %
    & \textsc{[el]} \small{\textsc{BitextEdit}} & 
     \selectlanguage{greek}
     Στη συνέχεια σταμάτησε, επιθεωρώντας την σουρεαλιστική σκηνή.\\ 
    \addlinespace[1.0em]
    \cmidrule{2-3}
    \addlinespace[1.0em]

    \editedside & \textsc{[en]} \small{\textsc{ccmatrix}} &
    \colorbox{red!10}{\hz Device installation error} is a frequent error. \\
    & \textsc{[el]} \small{\textsc{ccmatrix}} & 
    \selectlanguage{greek}
    Η ακατάλληλη φόρμα βιογραφικού, είναι ένα πολύ συχνό λάθος.  \\
     & \hspace{0.5em} $\lfloor$ \hspace{0.3em} \small{\textsc{gloss}}  &
    {\color{gray!80}
    \hspace{0.1em}
    \textit{An improper resume form, is a very frequent mistake.
    }}\\        %
    & \textsc{[en]} \small{\textsc{BitextEdit}} & 
    The inappropriate biographical form is a very frequent mistake. \\ 
    \addlinespace[1.0em]
    \cmidrule{2-3}
    \addlinespace[1.0em]
    & \textsc{[en]} \small{\textsc{ccmatrix}} & 
    I could work with a hospital specialist as a clinical assistant (as I have done). \\
    \editedside & \textsc{[el]} \small{\textsc{ccmatrix}} &
    \selectlanguage{greek}
    \colorbox{red!10}{\hz δούλευε ώς βοηθός ερευνητή παράλληλα με το διδακτορικό (όπως και εγώ)} \\
     & \hspace{0.5em} $\lfloor$ \hspace{0.3em} \small{\textsc{gloss}}  &
    {\color{gray!80}
    \hspace{0.1em}
    \textit{ They were working as an assistant researcher in parallel with their doctorate (as I have done).
    }}\\        %
    & \textsc{[el]} \small{\textsc{BitextEdit}} & 
    \selectlanguage{greek}
     Θα μπορούσα να δουλέψω με έναν ειδικό στο νοσοκομείο ως κλινικός βοηθός (όπως έχω κάνει).\\
    \addlinespace[1.0em]
    \cmidrule{2-3}
    \addlinespace[1.0em]
    \editedside & \textsc{[en]} \small{\textsc{ccmatrix}} &
    We should, however, always be striving to live a sustainable and \colorbox{red!10}{\hz kind life.} \\
    & \textsc{[el]} \small{\textsc{ccmatrix}} & 
    \selectlanguage{greek}
     Πάντα πρέπει να παλεύουμε για δίκαιη και βιώσιμη ειρήνη. \\
     & \hspace{0.5em} $\lfloor$ \hspace{0.3em} \small{\textsc{gloss}}  &
    {\color{gray!80}
    \hspace{0.1em}
    \textit{ We must always fight for a just and lasting peace.
    }}\\        %
    & \textsc{[en]} \small{\textsc{BitextEdit}} & 
     We must always fight for just and sustainable peace. \\
    \addlinespace[1.0em]
    \bottomrule
    \end{tabular}}
    \caption{Examples of CCMatrix bitexts along with refined sides generated by \textsc{BitextEdit}. \editedside \ \  denotes the side ([\textsc{el}] or [\textsc{en}]) that the model edits, while highlighted segments indicate the meaning mismatches in the original CCMatrix sentence that gets edited. Greek sentences are glossed to help understanding their meaning.}
    \label{tab:model_outputs}
\end{table*}


\begin{table*}[!t]
    \centering
    \scalebox{0.8}{
    \begin{tabular}{lllll}
    \toprule
    \textbf{Corpus} & \textbf{Version} & \textbf{License} & \textbf{Citation} & \textbf{Link} \\
    \hline
    CCmatrix & v2 & - & \citet{schwenk-etal-2021-ccmatrix} & \tiny{\url{https://data.statmt.org/cc-matrix/}} \\
    \textsc{flores} & v1 & CC-BY-SA & \citet{guzman-etal-2019-flores} & \tiny{\url{https://github.com/facebookresearch/flores}} \\
    OpenSubtitles & v2018 & - & \citet{lison-tiedemann-2016-opensubtitles2016} & \tiny{\url{https://opus.nlpl.eu/OpenSubtitles-v2018.php}}\\
    Tatoeba & v2 & CC–BY 2.0 FR & \citet{TIEDEMANN12.463} & \tiny{\url{https://opus.nlpl.eu/Tatoeba.php}}\\
    ParaCrawl &  v7.1 & Creative Commons CC0 & \citet{espla-etal-2019-paracrawl} & \tiny{\url{https://opus.nlpl.eu/ParaCrawl.php}}\\
    \bottomrule
    \end{tabular}}
    \caption{Additional documentation of scientific artifacts used in our paper.}
    \label{tab:artifacts_documentation}
\end{table*}

\section{Details on Scientific Artifacts}\label{sec:artifacts_details}
\paragraph{Statistics on Training Examples} Tables~\ref{tab:nmt_data_details} and \ref{tab:bitextedit_data_details} include detailed statistics on training and dev samples used to train each of the \textsc{nmt} and \textsc{BitextEdit} models discussed in the paper.
\begin{table}[!ht]
    \centering
    \scalebox{0.8}{
    \begin{tabular}{lrrrrr}
    \toprule
                       & \multicolumn{2}{c}{Training} & Dev & Test \\
    \hline
    Pair      & $|A|$ & $|A\cup B|$ &  & \\
    \textsc{en-oc}     & $242{,}982$ & $365{,}399$ & $997$ & $1{,}012$\\
    \textsc{it-oc}     & $309{,}703$ & $440{,}283$ & $997$ & $1{,}012$\\
    \textsc{en-be}     & $659{,}430$ & $3{,}944{,}412$ & $997$ & $1{,}012$\\
    \textsc{en-mr}     & $1{,}503{,}477$ & $3{,}611{,}336$ & $997$ & $1{,}012$\\
    \textsc{en-sw}     & $1{,}721{,}801$ & $2{,}641{,}234$ & $997$ & $1{,}012$\\
    \bottomrule
    \end{tabular}}
    \caption{Number of training/dev/test examples used to train \textsc{nmt} models in Table~\ref{tab:main_results}.}
    \label{tab:nmt_data_details}
\end{table}

\begin{table}[!ht]
    \centering
    \scalebox{0.8}{
    \begin{tabular}{lrrrr}
    \toprule
    Pair (src-tgt)      & All & Mined (src) & Mined (tgt) \\
    \hline
    \multicolumn{4}{c}{\textit{Training samples}} \\
    \hline
    \addlinespace[0.2cm]
    \textsc{en-oc}     & $3{,}822{,}800$  & $965{,}184$ & $946{,}216$\\
    \textsc{it-oc}     & $4{,}743{,}350$  & $1{,}228{,}328$ & $1{,}143{,}347$\\
    \textsc{en-be}     & $10{,}152{,}596$ & $2{,}637{,}575$ & $2{,}544{,}460$\\
    \textsc{en-mr}     & $17{,}764{,}241$ & $5{,}640{,}928$ & $5{,}991{,}336$\\ 
    \textsc{en-sw}     & $16{,}232{,}991$ & $6{,}734{,}355$ & $6{,}859{,}214$\\
    \addlinespace[0.2cm]
    \hline
    \multicolumn{4}{c}{\textit{Dev samples}} \\
    \hline   
    \addlinespace[0.2cm]
    \textsc{en-oc} & $15{,}908$ & $3{,}988$ & $3{,}966$ \\
    \textsc{it-oc} & $15{,}952$ & $3{,}988$ & $3{,}988$  \\
    \textsc{en-be} & $15{,}952$ & $3{,}988$ & $3{,}988$ \\
    \textsc{en-mr} & $15{,}952$ & $3{,}988$ & $3{,}988$ \\
    \textsc{en-sw} & $15{,}952$ & $3{,}988$ & $3{,}988$\\
    \bottomrule
    \end{tabular}}
    \caption{Number of training/dev examples used to train \textsc{BitextEdit} models in Table~\ref{tab:main_results}. The two last columns (i.e., \textit{mined}) include further statistics on the number of mined bitexts consumed by the \textit{edit-based reconstruction} loss; the rest of the training samples correspond to machine-translation samples upweighted to match the number of mined bitexts (i.e., equal contribution of two losses).}
    \label{tab:bitextedit_data_details}
\end{table}

\paragraph{License details} We use data derived from \textsc{opus} (\url{https://opus.nlpl.eu/}) corpora as summarized in Table~\ref{tab:artifacts_documentation}. All data are solely used for research purposes.

\begin{table}[!t]
    \centering
    \scalebox{0.75}{
    \begin{tabular}{lrrrr}
    \hline
    & \multicolumn{2}{c}{Original} & \multicolumn{2}{c}{Edited}  \\
    \hline
    & \multicolumn{4}{c}{\# \textit{Tokens}}\\
    & \textsc{src} & \textsc{tgt}   & \textsc{src} & \textsc{tgt} \\
      \textsc{en-oc}   & $3{,}591{,}876$ & $3{,}995{,}351$ & $3{,}601{,}179$ & $3{,}978{,}861$ \\ 
      \textsc{oc-it}   & $5{,}717{,}341$ & $5{,}496{,}704$ & $5{,}763{,}860$ & $5{,}428{,}767$ \\
      \textsc{be-en}   & $97{,}172{,}691$ & $43{,}007{,}326$ & $16{,}806{,}977$ & $19{,}002{,}607$ \\
      \textsc{en-mr}   & $36{,}468{,}349$ & $32{,}934{,}460$ & $36{,}411{,}035$ & $3{,}2830{,}479$ \\
      \textsc{en-sw}   & $4{,}1855{,}796$ & $40{,}666{,}513$ & $41{,}978{,}701$ & $40{,}472{,}724$ \\
      \hline
    & \multicolumn{4}{c}{\# \textit{Types}}\\
    \hline
    & \textsc{src} & \textsc{tgt}   & \textsc{src} & \textsc{tgt} \\
    \textsc{en-oc} & $165{,}310$ & $234{,}252$ & $169{,}191$ & $235{,}503$ \\
    \textsc{oc-it} & $277{,}397$ & $278{,}727$ & $292{,}357$ & $283{,}656$ \\
    \textsc{be-en} & $531{,}309$ & $526{,}289$ & $533{,}224$ & $381{,}666$ \\
    \textsc{en-mr} & $407{,}977$ & $956{,}589$ & $379{,}015$ & $922{,}184$\\
    \textsc{en-sw} & $414{,}873$ & $802{,}292$ & $409{,}0224$ & $791{,}853$ \\
    
        \hline
    & \multicolumn{4}{c}{\textit{Type-Token ratio}}\\
    \hline
    & \textsc{src} & \textsc{tgt}   & \textsc{src} & \textsc{tgt} \\
      \textsc{en-oc}   & $4.6\%$ & $5.9\%$ & $4.7\%$ & $5.9\%$ \\ 
      \textsc{oc-it}   & $5.9\%$  & $5.1\%$ & $5.1\%$ & $5.2\%$ \\
      \textsc{be-en}   & $0.5\%$ & $1.2\%$ & $3.2\%$ & $2.0\%$ \\
      \textsc{en-mr}   & $1.1\%$ & $2.9\%$ & $1.0\%$ & $2.8\%$ \\
      \textsc{en-sw}   & $2.0\%$ & $2.0\%$ & $1.0\%$ & $2.0\%$ \\
    
    \end{tabular}}
    \caption{Lexical characteristics of Original vs. Edited version of CCMatrix bitexts.}
    \label{tab:lexical_analysis}
\end{table}


\section{Compute Infrastructure \& Run time}\label{sec:infrastructure}
Each experiment runs on a single machine with 8 \textsc{gpu}s. \textsc{nmt} models require less than $3.5$ hours (e.g., \textsc{en-oc} on $A \cup B$ requires $\sim$ $20$ minutes to train). Similarly, \textsc{BitextEdit} models require less than $13.5$ hours to train (e.g., \textsc{en-oc} requires $\sim$ $5$ hours).
All models follow the transformer architecture detailed in Appendix~\ref{sec:fairseq_settings} with a total of $165$M parameters.
%

\section{Potential Risks}
\paragraph{Hallucination detection}
Our approach introduces synthetic samples (i.e., edited references that replace the originally human generated samples) that are later consumed as training instances by \textsc{nmt} models. One concern of using synthetic instances highlighted by recent work \cite{zhou-etal-2021-detecting}, is the generation of hallucinations (i.e., fluent text that is not tight to the source segment). To understand whether our method potentially contributes to the issue of hallucinations, one of the authors examined a small sample of $20$ outputs generated by three \textsc{nmt} models for \textsc{en}$\rightarrow$\textsc{el} translation: 
\begin{inparaenum}
\item a model trained only on $1$M of PoolA data;
\item a model trained on the concatenation of $1$M PoolA and $2$M PoolB data;
\item a model trained on the concatenation of $1$M PooA and $2$M edited PoolB data.
\end{inparaenum}
The \textsc{nmt} outputs are annotated labeled as: incomprehensible, faithful, or contains hallucinations following the protocol of  \citet{zhou-etal-2021-detecting}. All annotated instances are found to be faithful to the source. 

\paragraph{Lexical Richness} Synthetically generated data (e.g., machine-translated instances) are known to exhibit a decay in lexical richness when compared to human written texts \cite{vanmassenhove-etal-2019-lost}. To confirm that our approach does not potentially contribute to this issue, we report more detailed statistics on how the original and edited CCMatrix texts differ in terms of lexical features (i.e., \#tokens, \#types, and type-token ratio). As presented in Table~\ref{tab:lexical_analysis} the edited text does exhibit a decrease in the type-token ratio percentage when compared to the original one. 

\label{sec:appendix}

\end{document}